%% file: main.tex
\documentclass[dvipsnames,format=sigconf,anonymous=false,review=false,nonacm=true]{acmart}
\AtBeginDocument{%
  }

\setcopyright{acmlicensed}
\copyrightyear{2018}
\acmYear{2018}
\acmDOI{XXXXXXX.XXXXXXX}
\acmISBN{978-1-4503-XXXX-X/2018/06}

\usepackage{booktabs}
\usepackage{colortbl}
\usepackage{xcolor}
\usepackage{algorithm}
\usepackage{algorithmic}
\usepackage{tabularx}
\usepackage{multirow}
\usepackage{graphicx} 
\usepackage{makecell}

\definecolor{bestblue}{RGB}{170,190,230}        % 中蓝
\definecolor{lightblue}{RGB}{210,225,245}      % 浅蓝
\definecolor{verylightblue}{RGB}{235,242,252}  % 很浅蓝

\begin{document}

\title{A Learning-Based Cooperative Coevolution Framework for Heterogeneous Large-Scale  Global Optimization}

\author{Wenjie Qiu}
\email{wukongqwj@gmail.com}
\orcid{0009-0000-1965-8863}
\affiliation{%
  \institution{South China University of Technology}
  \city{Guangzhou}
  \state{Guangdong}
  \country{China}
}

\author{Zixin Wang}
\email{joxin64@gmail.com}
\orcid{0009-0009-2453-7682}
\affiliation{%
  \institution{South China University of Technology}
  \city{Guangzhou}
  \state{Guangdong}
  \country{China}
}

\author{Hongyu Fang}
\email{hongyufang969@gmail.com}
\orcid{0009-0007-1953-7261}
\affiliation{%
  \institution{South China University of Technology}
  \city{Guangzhou}
  \state{Guangdong}
  \country{China}
}

\author{Zeyuan Ma}
\email{scut.crazynicolas@gmail.com}
\orcid{0000-0001-6216-9379}
\affiliation{%
  \institution{South China University of Technology}
  \city{Guangzhou}
  \state{Guangdong}
  \country{China}
}

\author{Yue-Jiao Gong}
\email{gongyuejiao@gmail.com}
\authornote{Corresponding Author}
\orcid{0000-0002-5648-1160}
\affiliation{%
  \institution{South China University of Technology}
  \city{Guangzhou}
  \state{Guangdong}
  \country{China}
}

% \affiliation{%


\begin{abstract}
Cooperative Coevolution (CC) effectively addresses Large-Scale Global Optimization (LSGO) via decomposition but struggles with the emerging class of Heterogeneous LSGO (H-LSGO) problems arising from real-world applications, where subproblems exhibit diverse dimensions and distinct landscapes. The prevailing CC paradigm, relying on a fixed low-dimensional optimizer, often fails to navigate this heterogeneity. To address this limitation, we propose the Learning-Based Heterogeneous Cooperative Coevolution Framework (LH-CC). By formulating the optimization process as a Markov Decision Process, LH-CC employs a meta-agent to adaptively select the most suitable optimizer for each subproblem. We also introduce a flexible benchmark suite to generate diverse H-LSGO problem instances. Extensive experiments on 3000-dimensional problems with complex coupling relationships demonstrate that LH-CC achieves superior solution quality and computational efficiency compared to state-of-the-art baselines. Furthermore, the framework exhibits robust generalization across varying problem instances, optimization horizons, and optimizers. Our findings reveal that dynamic optimizer selection is a pivotal strategy for solving complex H-LSGO problems.

\end{abstract}





\maketitle

\section{Introduction}

\begin{figure}[h]
  \centering
  \includegraphics[width=\linewidth]{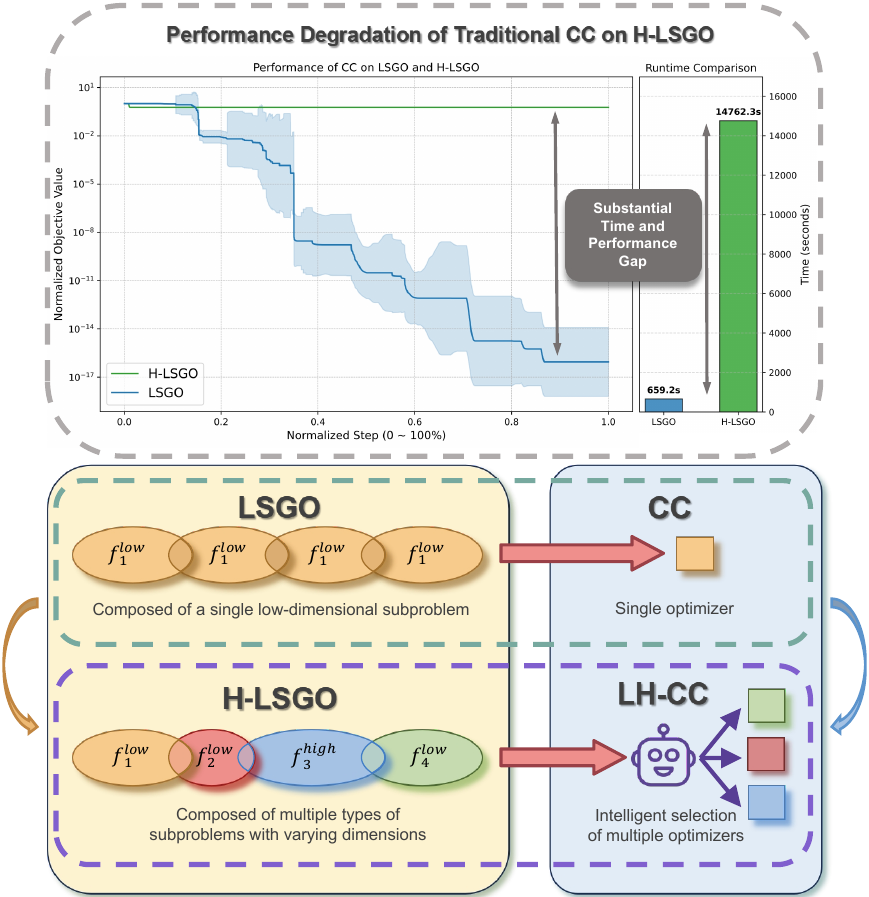}
  \caption{Motivation: Shifts in problem paradigms drive corresponding shifts in optimization paradigms.}
  \label{figure:idea}
\end{figure}

Large-Scale Global Optimization (LSGO), which involves black-box problems with dimensions exceeding 1000, remains one of the most formidable challenges in the field of Evolutionary Computation \cite{omidvar2021review}. Such problems are ubiquitous in real-world applications, ranging from neuroevolution \cite{galvan2021neuroevolution} to the design of electric aircraft \cite{ha2019large} and wind turbines \cite{thomas2017improving}. The primary challenge of LSGO stems from the "curse of dimensionality," where the search space expands exponentially with increasing dimensions, rendering conventional algorithms designed for low-dimensional problems ineffective \cite{yao2025coevolutionary}. To address LSGO, Cooperative Coevolution (CC) is a widely adopted framework \cite{ma2018survey,gong2023offline}. Inspired by the divide-and-conquer paradigm, CC decomposes a high-dimensional problem into multiple low-dimensional subspaces for independent optimization. The resulting sub-solutions are then concatenated to form a complete candidate solution \cite{qiu2025novel}. 

Based on an analysis of diverse practical applications \cite{ibrahimov2012evolutionary,song2019divide}, it is observed that the optimization problems therein typically exhibit a structure composed of multiple homogeneous subproblems. In such cases, the distinction between subproblems lies solely in their specific dimensions, yet they remain fundamentally low-dimensional. Furthermore, their interactions are predominantly limited to simplistic sequential coupling. Such structural properties were formally integrated into the CEC2013LSGO~\cite{li2013benchmark} suite, serving as the classic  benchmark paradigm that has guided the development of most state-of-the-art LSGO algorithms~\cite{omidvar2017dg2,kumar2022efficient,komarnicki2024overlapping}. Under this setting, the main challenge usually lies in decomposition~\cite{liu2024large,yao2025cooperative}, while a single low-dimensional optimizer is often sufficient after decomposition~\cite{liu2025evolutionary}.

However, as research progresses, increasingly complex problems have emerged. A prime example is satellite design, which involves seven modules, i.e., orbit dynamics, attitude dynamics, cell illumination, solar power, temperature, energy storage, and communication \cite{hwang2014large}. 
% Each module presents a unique optimization problem with varying degrees of inter-module coupling.
These problems exhibit pronounced subproblem heterogeneity. Specifically, each problem comprises multiple heterogeneous subproblems, whose functional forms and dimensionality can vary substantially across subproblems. For the sake of distinction, we refer to this class of problems as Heterogeneous Large-Scale Global Optimization (H-LSGO). Preliminary studies show that, although advanced decomposition methods remain effective for identifying variable interactions, overall optimization performance degrades markedly and runtime increases sharply on H-LSGO \cite{xu2023large}. This performance and time gap primarily stems from the fundamental mismatch between the inherent heterogeneity of subproblems and the traditional paradigm of employing a singular, low-dimensional optimizer.
% A pivotal reason for this decline is the limitation of employing a singular, low-dimensional optimizer (e.g., CMA-ES \cite{auger2012tutorial}, SHADE \cite{tanabe2013success}, CLPSO \cite{liang2006comprehensive}) for all subproblems. % 这里要改写一下逻辑，应该是由于问题的异质性和优化器的维度不匹配，但现在仍然是用单一的
% This hindrance manifests in two ways: 1) Heterogeneous Requirements: different subproblems necessitate distinct exploration-exploitation balances, requiring diverse optimization strategies; 2) Dimensionality Mismatch: for subproblems that are no longer low-dimensional, optimizers specifically tailored for small-scale spaces often struggle to maintain efficacy.

% 问题范式的改变推动着优化范式的改变。
This shift in the problem paradigm inevitably necessitates a corresponding transition in the optimization paradigm. Meta-Black-Box Optimization (MetaBBO) is an emerging paradigm that aims to automate the design of optimization algorithms \cite{ma2025metabox,ma2025toward,chen2025metade}. Many existing MetaBBO approaches have shown on low-dimensional problems that automatically designed algorithms can be highly efficient and capable of outperforming manually crafted counterparts, and generalize well across problems \cite{guo2025designx,guo2025advancing,xu2025autoep, song2024reinforcement}. Compared with low-dimensional problems, H-LSGO problems are more complex, and expert knowledge for designing effective algorithms here is considerably more scarce.
Motivated by the reduced reliance on expert knowledge enabled by data-driven MetaBBO, we propose LH-CC, a \textbf{L}earning-based \textbf{H}eterogeneous \textbf{C}ooperative \textbf{C}oevolution framework, for automated algorithm design to solve complex H-LSGO problems \cite{cenikj2024survey, cenikj2024transoptas}. Figure \ref{figure:idea} shows the core motivation of LH-CC and the main contributions of this paper are as follows:
\begin{itemize}
    \item \textbf{LH-CC stands as the pioneering CC framework specifically designed to address H-LSGO problems} : It extends the existing CC horizon by implementing a learning-based dynamic optimizer selection mechanism, effectively countering the novel H-LSGO problem paradigm through a paradigm shift in optimization. By formulating the optimization process as a Markov Decision Process, we meticulously design a suite of tailored states, actions, and network.
    \item \textbf{The First Benchmark for Automated H-LSGO Instance Generation} : Existing research on H-LSGO predominantly relies on manually designed benchmarks, which are constrained by limited diversity in both problems and structural characteristics. To address this, we introduce a user-friendly benchmark designed for the automated generation of H-LSGO problems. This tool enables the efficient and flexible creation of tailored problem instances, effectively alleviating the current scarcity of diverse LSGO test benchmarks. 
    \item \textbf{LH-CC demonstrates exceptional competitiveness in both optimization quality and efficiency, alongside robust generalizability} :  Based on our proposed benchmark, we constructed a suite of 3000-dimensional H-LSGO problems featuring complex coupling relationships and conducted extensive experiments. The results reveal that LH-CC exhibits a significant competitive advantage, confirming its effectiveness in extending the prevailing CC optimization paradigm. Moreover, the framework possesses strong generalizability, capable of seamless transfer across diverse problem instances and optimizers. Finally, through ablation studies, we underscore that dynamic optimizer selection is a pivotal strategy for addressing H-LSGO challenges.
\end{itemize}

The remainder of this paper is organized as follows: Section \ref{sec:background} reviews related work, followed by the details of the LH-CC architecture in Section \ref{sec:method}. Section \ref{sec:hlsgo} describes the Auto-H-LSGO benchmark. Section \ref{sec:workflow} details the overall workflow and training procedure of LH-CC. Section \ref{sec:experiment} presents experimental results and insights, while Section \ref{sec:conclusion} concludes the paper.

% The remainder of this paper is organized as follows: Section \ref{sec:background} reviews related work, followed by the details of the LH-CC architecture in Section \ref{sec:method}. Section \ref{sec:experiment} presents experimental results and insights, while Section \ref{sec:conclusion} concludes the paper.

\section{Background}
\label{sec:background}
\subsection{LSGO and CC}
\label{sec:lsgo-cc}
Large-Scale Global Optimization (LSGO) usually refers to black-box problems with more than 1000 variables. The CEC2013LSGO~\cite{li2013benchmark} suite,a standard benchmark in this field, mainly contains problems composed of several homogeneous subproblems. Taking F9, a representative problem, as a case in point:
\begin{equation}
\label{F9}
    F_{9}(\mathbf{z})=\sum_{i=1}^{20} w_{i} f_{\text {elliptic }}\left(\mathbf{z}_{i}\right)
\end{equation}
Here, $w_{i}$ denotes the weight coefficient of the subproblems. It can be observed that F9 is formulated as the sum of 20 Elliptic-based subproblems. Their dimensionalities take three values (25, 50, and 100), all within the low-dimensional regime.

In practice, structural information such as the dimensionalities of the underlying subproblems is typically unknown. Many advanced CC decomposition strategies (e.g., DG2~\cite{omidvar2017dg2}), based on variable-interaction analysis, can decompose the search space into independent subspaces corresponding to individual subproblems. Subsequently, a single low-dimensional optimizer (e.g., CMA-ES~\cite{auger2012tutorial}) can be applied sequentially to each subspace to achieve satisfactory optimization performance. However, as research has progressed, more complex optimization problems have emerged, such as the example below: 
\begin{equation}
\label{FH}
    F(\mathbf{z})=\sum_{i=1}^{5} w_{i} f_{\text {elliptic}}\left(\mathbf{z}_{i}\right)+\sum_{i=1}^{5} w_{i} f_{\text {schwefel}}\left(\mathbf{z}_{i}\right)+\sum_{i=1}^{10} w_{i} f_{\text {ackley}}\left(\mathbf{z}_{i}\right)
\end{equation}
Comparing Equation~\ref{F9} and~\ref{FH}, a salient difference is that the latter is no longer composed of a single type of subproblem. Moreover, individual subproblems may themselves be high-dimensional, e.g., with dimensionalities of 500 or 1000. This heterogeneity at the subproblem level clearly distinguishes it from conventional LSGO problems. We therefore refer to these problems as Heterogeneous Large-Scale Global Optimization (H-LSGO).
The conventional paradigm of applying a single low-dimensional optimizer to solve H-LSGO problems suffers from clear limitations in both optimization performance and runtime efficiency. This bottleneck, in turn, motivates a shift in the optimization paradigm.

\begin{table*}[t]
% \small % 如果空间依然紧凑，可以尝试 \footnotesize
\centering
\caption{State Features in LH-CC}
\label{tab:state-feature-summary}

% ---------------------------------------------------------
% % 调优参数
\renewcommand{\arraystretch}{1.2} % 稍微增加行高，防止公式挤在一起
% \setlength{\tabcolsep}{5pt}      % 适当间距
% ---------------------------------------------------------
\resizebox{\linewidth}{!}{
\begin{tabular}{cccl}
% {\textwidth}{l p{0.1cm} >{\centering\arraybackslash}p{3.9cm} X} % 增加第二列宽度到4.8cm
\hline
\textbf{Feature} & \textbf{Index} & \textbf{Formula} & \textbf{Explan} \\
\hline

\multirow{3}{*}{$\mathcal{S}_{\text{problem}}$} 
& 1 & $(D_k / 500)^{0.4}$ & The relative size of the subproblem dimension $D_k$. \\
& 2 & $\mathcal{I}(\left| \left\{ i \mid \sum_{j=1}^{D} \Theta[i,j] = 1 \right\} \right| = D)$ & The separability indicator, derived from the DSM $\Theta$, where the dependency of variables is reflected. \\
& 3 & $\left|\bigcup_{i=1}^{|S|}\bigcup_{j\neq i}^{|S|}(\Omega_i\cap\Omega_j)\right|/D$ & The ratio of shared variables among all subcomponent pairs ($\Omega_i \cap \Omega_j$) to the number of variables $D$.

 \\

\hline

\multirow{5}{*}{$\mathcal{S}_{\text{pop}}$} 
& 1 & $d=\frac{1}{N(N-1)}\!\sum_{i\neq j}\!\|x_i-x_j\|$ & The average pairwise distance among all individuals in the population. \\
& 2 & $d_{\text{top}} - d$ & The difference between the average pairwise distance $d_{\mathrm{top}}$ of the top (10\%) individuals and $d$.\\
& 3 & $\frac{1}{NS}\sum_{s=1}^S\sum_{i=1}^N\mathcal{I}\bigl(|c_i-c_i^s|<\epsilon\bigr)$ & The proportion of sampled individual costs $c_i^s$ that are close to the original cost $c_i$ in a tolerance $\epsilon$. \\
& 4 & $\frac{1}{N}\sum_{i=1}^N \mathcal{I}(\sum_{s=1}^S \mathcal{I}(c^s_i < c_i) = 0)$ & The proportion of individuals whose sampled costs $c_i^s$ never improve upon the current cost $c_i$. \\
& 5 & $\frac{1}{N}\sum_{i=1}^N \mathcal{I}(\sum_{s=1}^S \mathcal{I}(c^s_i > c_i) < S)$ & The proportion of individuals whose sampled costs $c_i^s$ do not consistently exceed the current cost $c_i$. \\

\hline

\multirow{6}{*}{$\mathcal{S}_{\text{progress}}$} 
& 1 & $\mathrm{FEs} / \mathrm{MaxFEs}$ & The ratio of  consumed function evaluations FEs to the total evaluations MaxFEs. \\
& 2 & $(\log_{10}(c_t^\ast + \delta_{\mathrm{f}}) / \log_{10} ( c_{0}^* + \delta_{\mathrm{f}}))^2$ & The solution quality, ratio of two logged, offset ($\delta_{\mathrm{f}}$) values, current-best cost ${c}_t^{*}$ and initial cost ${c}_0^*$. \\
& 3 & $(c_t^\ast / c_{t-1}^\ast)^{8}$ & The relative improvement ratio of the global best cost $c^{\ast}$ between two consecutive steps $t$ and $t-1$. \\
& 4 & $(c_{t}^{\ast(k)} / c_{t-1}^{\ast(k)})^{8}$ & The relative improvement ratio of the group-specific best cost for a subproblem $k$. \\
& 5 & $\mathrm{FEs}_l / \mathrm{FEs}_{\text{total}}$ & The usage proportion in terms of function evaluations, maintained per optimizer $l$. \\
& 6 & $\Delta\log_{10}(c^{(l)}) / \max(\log_{10}( c_0^*) - \log_{10}( c_t^\ast) , 0.1)$ & The ratio of optimizer-specific log-scale improvement $\Delta \log_{10} c^{(l)}$ to the total for each optimizer $l$. \\
\hline
\end{tabular}
}
\end{table*}

\subsection{MetaBBO}

Meta-Black-Box Optimization (MetaBBO) is an emerging paradigm for automating the design of optimization algorithms ~\cite{ma2025metabox,ma2025toward,li2024bridging,song2024reinforcement,qiu2026automated}, and is typically formulated as a Markov decision process (MDP).
% frames the design of optimization algorithms as a "learning to learn" problem, 
% This paradigm allows for the automated discovery of search strategies by interacting with optimization environments.
% Formally, the MetaBBO process is defined by a quadruple $(\mathcal{S}, \mathcal{A}, \mathcal{T}, \mathcal{R})$:  
% A state $s_t \in \mathcal{S}$ characterizes the status of the optimization process at time step $t$, such as population statistics or landscape features;  
% An action $a_t \in \mathcal{A}$ represents a strategy decision, like selecting search operators or adjusting hyper-parameters;  
% $\mathcal{T}(s_{t+1}|s_t, a_t)$ models the optimization dynamics;  
% $\mathcal{R}(s_t, a_t)$ gives the objective improvement between steps.
Formally, the MetaBBO process is formulated as a tuple $(\mathcal{S}, \mathcal{A}, \mathcal{T}, R)$. Specifically, the meta-agent $\pi_{\theta}$, at each optimization step $t$, selects the most suitable action $a_t$ from a pool of actions $\mathcal{A}$ based on the current problem $F_i$ optimization state info $s_i^t \in \mathcal{S}$, with the aim of achieving better optimization performance. The environment then transitions to a new state $s_{t+1}$ according to the transition probability $\mathcal{T}(s_{t+1} \mid s_t, a_t)$. The reward function $R: \mathcal{S} \times \mathcal{A} \rightarrow \mathbb{R}$ provides feedback in the form of a scalar reward $r_t$ corresponding to the state-action pair.

% the state $s_t \in \mathcal{S}$ encapsulates the optimization states at step $t$. The action $a_t \in \mathcal{A}$ corresponds to a algorithm design strategy. Furthermore, $\mathcal{T}(s_{t+1}|s_t, a_t)$ models the underlying optimization dynamics, while $\mathcal{R}(s_t, a_t)$ quantifies the improvement in the objective function between consecutive steps.
% Technically, MetaBBO operates within a two-level optimization framework. The \textit{lower-level} (optimization layer) executes selected algorithms on specific problem instances, while the \textit{upper-level} (meta-learning layer) optimizes the parameters $\theta$ of a meta-policy $\pi_\theta(a_t \mid s_t)$.
Assuming a finite decision horizon of $T$ steps, the interaction between the agent and the environment yields a trajectory $\tau := (s_0, a_0, s_1, \dots, s_T)$.
The learning objective of the meta-agent is to maximize the optimization performance $J(\theta)$ of the algorithm design strategy over the target problem distribution $\mathcal{P}$ (which is often discretized by sampling $N$ problem instances $F_i$ from the distribution to form a problem set $\Upsilon = \{F_1,\dots,F_N \}$):
\begin{equation}
\label{eq:metabbo_obj}
    \begin{aligned}
        J(\theta) &= \mathbb{E}_{f_i \in \mathcal{P}} [\operatorname{perf}(\mathcal{S}, \mathcal{A})] \\
        &\approx \frac{1}{N} \sum_{i=1}^{N} \sum_{t=1}^{T} \operatorname{perf}\left(s_i^t, a_t = \pi_{\theta}(s_i^t)|F_{i}\right)
    \end{aligned}
\end{equation}
where $\operatorname{perf}(\cdot)$ is the performance evaluation function, used to quantify the effectiveness of the selections. 
% The objective is to find parameters $\theta$ that maximize the expected cumulative reward over training problems $\mathcal{P}$:
% \begin{equation}
% \label{eq:metabbo_obj}
%     \theta^* = \arg\max_{\theta} 
%     \mathbb{E}_{p \sim \mathcal{P}} \left[ \sum_{t=1}^{T} r_t \right],
% \end{equation}
% where $T$ is the maximum number of decision steps in one optimization episode.

Extensive research indicates that MetaBBO is capable of efficient automated algorithm design even under limited expert knowledge, yielding solutions that surpass manually crafted counterparts in both performance and generalizability~\cite{du2025metablackboxoptimizationbispacelandscape, ma2025accurate, van2025llamea}.

% \begin{figure*}[h]
%   \centering
%   \includegraphics[width=0.9\linewidth]{assets/ccrldas_structure.drawio.png}
%   \caption{Workflow of AS, RL-DAS, and proposed LH-CC}
%     \label{figure:as-comparasion}
%   \Description{Workflow of AS, RL-DAS, and LH-CC}
% \end{figure*}
    
\section{Methodology}
\label{sec:method}

To enable LH-CC to flexibly adapt to the diverse heterogeneous subproblems in H-LSGO, we model its optimization process as a MDP. The goal of LH-CC is to train an optimal meta-agent $\pi^*$ via deep reinforcement learning such that, at each decision step $t$, the agent selects the optimizer $a_t$ that best suits each heterogeneous subproblem $f_i$, thereby maximizing $J(\theta)$ over the target problem distribution $\mathcal{P}$ (Assume that $F_j$ comprises $M_j$ subproblems):
\begin{equation}
\label{eq:objective}
\begin{aligned} 
&\pi^{*} =\underset{\pi_{\theta} \in \Pi}{\arg \max } J(\theta) \\ & \approx \underset{\pi_{\theta} \in \Pi}{\arg \max } \frac{1}{N} \sum_{j=1}^{N} \sum_{i=1}^{M_j} \sum_{t=1}^{T} \gamma^{t-1} \operatorname{perf}\left(s_{i}^{t}, a_{t}=\pi_{\theta}\left(s_{i}^{t}\right) \mid f_{i}\right)
\end{aligned}
\end{equation}
where $\gamma$ discounts future rewards. Parameterized by a neural network $\pi_\theta$, the policy can be optimized via gradient-based methods (e.g., Proximal Policy Optimization (PPO) \cite{schulman2017proximal}) to approximate the optimal agent $\pi^*$.
% We formulate the adaptive selection of optimizers for heterogeneous subproblems in H-LSGO as a Markov Decision Process (MDP). The agent is represented by a parameterized policy $\pi_\theta$, which interacts with the optimization environment to maximize cumulative reward. The goal of LH-CC is to learn parameters $\theta^*$ through deep reinforcement learning such that, at each decision step $t$, the policy selects an action $a_t$ (i.e., an optimizer) for each decomposed subproblem.

% In practice, the expectation in Eq.~(\ref{eq:metabbo_obj}) is approximated using $M$ sampled training problems:
% \begin{equation}
% \label{eq:objective}
% \begin{aligned} \pi^{*} & =\underset{\pi_{\theta} \in \Pi}{\arg \max } J(\theta) \\ & \approx \underset{\pi_{\theta} \in \Pi}{\arg \max } \frac{1}{N} \sum_{i=1}^{N} \sum_{t=1}^{T} \gamma^{t-1} \operatorname{pcrf}\left(s_{i}^{t}, a_{t}=\pi_{\theta}\left(s_{i}^{t}\right) \mid f_{i}\right)\end{aligned}
% \end{equation}
% where $r_t^{(i)}$ denotes the reward obtained at step $t$ on the $i$-th training problem.

% We employ Proximal Policy Optimization (PPO) \cite{ppo} to optimize the parameters $\theta$ of the neural network $\pi_\theta$, as it provides the stability and sample efficiency required for high-dimensional black-box optimization.

\subsection{MDP Formulation}

\subsubsection{Action}
\label{sec:method:action}
The mismatch between optimizers and the dimensionality or type of subproblems is a primary cause of performance degradation in CC for H-LSGO. Consequently, to adapt to diverse heterogeneous subproblems, the action pool $\mathcal{A}$ of LH-CC consists of both high-dimensional $\mathcal{A}_\text{high}$ and low-dimensional optimizers $\mathcal{A}_\text{low}$. Given $L$ candidate optimizers, the composite action space is defined as:
\begin{equation}
\mathcal{A} = \mathcal{A}_\text{high} \times \mathcal{A}_\text{low}, \mathcal{A}_\text{high} \in \{1,\dots,j \}, \mathcal{A}_\text{low} \in \{j+1, \dots, L\}.
    % \mathcal{A}=\{ \mathcal{A}_\text{high}=\{a_1,\dots,a_i\}, \mathcal{A}_\text{low}=\{a_{i+1} \dots a_{L} \}   \}
\end{equation}
This enables the meta-agent to select the optimizer best suited for the specific problem from the dimensionally compatible candidates.

\subsubsection{State}
\label{sec:method:state}
The state $\mathcal{S}$ comprises three categories of features: problem $\mathcal{S}_\text{problem} \in \mathbb{R}^3$, population $\mathcal{S}_\text{pop} \in \mathbb{R}^5$ and optimization-progress $\mathcal{S}_\text{progress} \in \mathbb{R}^{4+2\times L}$. They integrate static ($\mathcal{S}_\text{problem}$) and dynamic features ($\mathcal{S}_\text{pop}$, $\mathcal{S}_\text{progress}$), as well as local and global attributes, to provide the meta-agent $\pi_\theta$ with comprehensive optimization state information, as summarized in Table~\ref{tab:state-feature-summary}:

\begin{figure*}[h]
  \centering
  \includegraphics[width=0.9\linewidth]{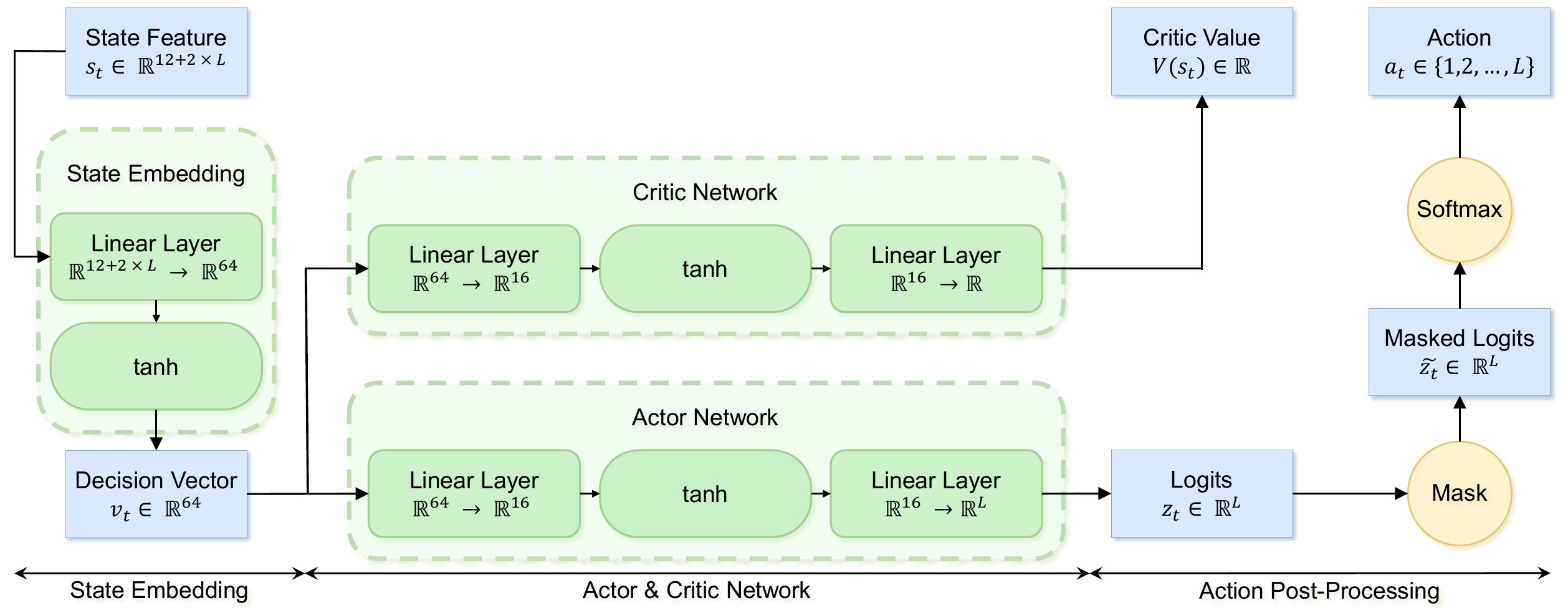}
  \caption{The network of LH-CC}
  \label{figure:network}
\end{figure*}

% is represented by a set of features extracted from both the population and the 
% problem landscape. We categorize all features into three groups:
For heterogeneous optimization, identifying subproblem characteristics is paramount. Accordingly, we first design $\mathcal{S}_\text{problem}$ to encode intrinsic problem attributes, including dimensionality, separability, and overlap ratio. Furthermore, to characterize the diverse fitness landscapes exhibited by different subproblems, we extract dynamic landscape information from the population distribution, denoted as $\mathcal{S}_\text{pop}$. This includes metrics such as the Non-improvable Ratio (NI) and Average Neutral Ratio (ANR). Finally, to monitor the optimization trajectory, we introduce $\mathcal{S}_\text{progress}$, which tracks iteration counts, computational budget consumption, and relative performance improvements.

% 1)~\emph{Optimization-Progress Features} $S_\text{progress}$: are extracted by tracking  iteration steps. They measure algorithmic progress and evaluation-budget consumption, as well as improvement relative to former steps.
% 2)~\emph{Population Features} $S_\text{pop}$: are extracted from population obtained in the process of optimizing of the algorithms. They quantify current population structure, convergence status, and evolvability.
% 3)~\emph{Problem Features} $S_\text{problem}$: describe inherent properties of the problem instance, including separability, overlap and so on.

The detailed definition and calculation of states are provided in Appendix \ref{sec:feature}.

% The action $a_t$ selects an optimizer from $\mathcal{O}$, which is partitioned into low-dimensional ($\mathcal{O}_{low}$) and high-dimensional ($\mathcal{O}_{high}$) groups. To prevent ineffective allocation, we employ a masking process:

% \begin{enumerate}
%     \item \textbf{Logits Generation:} The Actor network outputs raw logits $\mathbf{z}_t \in \mathbb{R}^L$.
%     \item \textbf{Mask Construction:} Using the dimension feature $f_{size}$ and a threshold $\tau$, the mask $m_t \in \{0,1\}^L$ is defined as:
%     \begin{equation}
% m_{t,l} =
% \begin{cases}
% 0, & \text{if } f_{\text{size}} > \tau \text{ and } o_l \in \mathcal{O}_{\text{low}},\\
% 1, & \text{otherwise},
% \end{cases}
% \quad l = 1,\dots,L.
%     \end{equation}
%     \item \textbf{Policy Transformation:} The masked policy $\pi$ is derived via a constrained Softmax:
%     \begin{equation}
% \pi_\theta(a_t = l \mid s_t)
% =
% \frac{\exp\!\big(z_{t,l} + \ln(m_{t,l}+\varepsilon)\big)}
% {\sum_{j=1}^{L} \exp\!\big(z_{t,j} + \ln(m_{t,j}+\varepsilon)\big)},
%     \end{equation}
%     where $\epsilon$ is a small constant to ensure numerical stability.
% \end{enumerate}

\subsubsection{Reward}
\label{sec:reward}

% To encourage the agent to continuously optimize the objective function across varying magnitudes,
To address the significant fitness-scale disparities in H-LSGO,
we design a reward mechanism based on the improvement of the global best cost.
 % , denoted as $c_t^\ast$. We define the initial best cost as the reference scale $c_s = c_0^\ast$, and its logarithmic transform $\log_{10}(c_s)$.
The core metric is the logarithmic reduction in global best cost $c_t^\ast$ at step $t$:
\begin{equation}
\Delta_{\log}= \log_{10}(c_{t-1}^{*})-\log_{10}(c_{t}^{*}),    
\end{equation}
which captures the order-of-magnitude progress. We construct the reward by normalizing the improvement against the reference scale derived from the initial best cost $c_0^*$, applying exponential shaping to enhance sensitivity to small progress. The reward $r_t$ at step $t$ is defined as:
\begin{equation}
r_{t}=\left(\frac{\Delta_{\log} + \delta_{\mathrm{r}}}{\log_{10}(c_0^*) + \delta_{\mathrm{r}}}\right)^{0.5}. %,\quad\mathrm{where}\;\eta=0.5.
\end{equation}
Here, $\delta_{\mathrm{r}}$ serves as an adaptive offset operating entirely in the logarithmic domain to stabilize the reward signal:
\begin{equation}
\delta_{\mathrm{r}}=\operatorname*{max}\left(1.5-\Delta_{\log},\;1.5-\log_{10}(c_0^*),\;0\right).
\end{equation}

\subsection{Network Design}

The proposed LH-CC framework adopts an Actor-Critic architecture to parameterize the meta-agent $\pi_\theta$. As illustrated in Figure~\ref{figure:network}, the network comprises three key modules: the \emph{State Embedding}, the \emph{Actor Network} for optimizer selection, and the \emph{Critic Network} for value estimation.

% At each decision step and for each subcomponent, the Actor produces a probability distribution over candidate optimizers, while the Critic estimates the expected return of the current optimization state.
% Both networks share the same state embedding module to ensure consistent representation learning.

\subsubsection{State Embedding}

Let $s_t \in \mathbb{R}^{12+2\times L}$ denote the state vector at step $t$, as defined in Section~\ref{sec:method:state}. To extract a compact and stable feature representation, we employ a lightweight state embedding module shared by both the Actor and Critic.

This embedding is implemented as a single fully connected layer:
\begin{equation}
\mathbf{e}_{t}=\operatorname{tanh}\left(W_{e}s_{t}+b_{e}\right),
\end{equation}
where $W_{e}\in\mathbb{R}^{64\times(12+2\times L)}$ and $b_{e}\in\mathbb{R}^{64}$ represent the trainable weights and biases, and $\mathbf{e}_{t}\in\mathbb{R}^{64}$ is the resulting state embedding. 
The $\tanh$ activation is utilized to bound feature magnitudes, thereby enhancing training stability. 
% Note that all subsequent matrices $W_{\{\cdot\}}$ and vectors $b_{\{\cdot\}}$ denote trainable parameters, with specific dimensions omitted for brevity unless necessary. 
The embedding $\mathbf{e}_{t}$ serves as the common input to both downstream networks.
% where $W_{e}\in\mathbb{R}^{(12+2\times L)\times64}$ and $b_{e}\in\mathbb{R}^{64}$ denote the trainable weights and biases of the embedding layer, respectively; $\mathbf{e}_{t}\in\mathbb{R}^{64}$ is the embedded state representation. The hyperbolic tangent activation is used to bound feature magnitudes and improve training stability.

% All subsequent weight matrices $W_{\{\cdot\}}$ and bias vectors $b_{\{\cdot\}}$ in the Actor and Critic networks are similarly trainable parameters, with their dimensions specified in context where necessary.

% This shared embedding serves as the input to both the Actor and the Critic networks.

\subsubsection{Actor Network}

The Actor network $\pi_\theta(a_t \mid s_t)$ outputs a categorical distribution over the candidate optimizer pool $\mathcal{A}$. Given the state embedding $\mathbf{e}_t$, the Actor first computes the unnormalized logits $\mathbf{z}_t \in \mathbb{R}^{L}$ via a two-layer perceptron:
\begin{equation}
\mathbf{z}_t = W_2 \tanh(W_1 \mathbf{e}_t + b_1) + b_2,
\label{eq:actor-logits}
\end{equation}
where $W_1 \in \mathbb{R}^{16 \times 64}$ and $W_2 \in \mathbb{R}^{L \times 16}$ are weight matrices, and $b_1, b_2$ are bias vectors. To prevent the application of computationally expensive low-dimensional optimizers ($\mathcal{A}_{\text{low}} \subset \mathcal{A}$) to high-dimensional subproblems—we employ an action masking mechanism. Let $s_{dim}$ denote the first component of $s_t$, representing the subspace dimensionality. A validity mask $\mathbf{m}_t \in \{-\infty, 0\}^L$ is constructed as follows:
\begin{equation}
m_{t,i} = 
\begin{cases}
-\infty, & \text{if } s_{\text{dim}} > 0.5 \text{ and } i \in \mathcal{A}_{\text{low}} \\
0,       & \text{otherwise}
\end{cases},
\end{equation}
The masked logits are then obtained by $\tilde{\mathbf{z}}_t = \mathbf{z}_t + \mathbf{m}_t$. The policy distribution is derived by applying the softmax function to the masked logits, ensuring zero probability for invalid actions:
\begin{equation}
\pi_\theta(a_t = i \mid s_t) = \frac{e^{\tilde{z}_{t,i}}}{\sum_{j=1}^{L} e^{\tilde{z}_{t,j}}}.
\end{equation}
Finally, the action $a_t$ is sampled from the policy distribution.

\subsubsection{Critic Network}

The Critic network approximates the state value function $V_\phi(s_t)$, which serves as the baseline for advantage computation. Sharing the state embedding $\mathbf{e}_t$ with the Actor, the Critic maps it to a scalar value via a two-layer perceptron:
\begin{equation}
V_\phi(s_t) = W_c^{(2)} \tanh(W_c^{(1)} \mathbf{e}_t + b_c^{(1)}) + b_c^{(2)},
\label{eq:critic}
\end{equation}
where $W_c^{(1)} \in \mathbb{R}^{16 \times 64}$ and $W_c^{(2)} \in \mathbb{R}^{1 \times 16}$ are the weight matrices. The Critic is optimized jointly with the Actor within the PPO framework to minimize the value estimation error.

% The Critic network estimates the state value function $V_\phi(s_t)$, which serves as a baseline for advantage computation.
% It shares the same state embedding $\mathbf{e}_t$ as the Actor, and maps it to a scalar value through a two-layer perceptron:
% \begin{equation}
%     V_\phi(s_t) = W_c^{(2)} \tanh(W_c^{(1)} \mathbf{e}_t + b_c^{(1)}) + b_c^{(2)},
%     \label{eq:critic}
% \end{equation}
% where $W_c^{(1)} \in \mathbb{R}^{64 \times 16}$ and $W_c^{(2)} \in \mathbb{R}^{16 \times 1}$.

% The Critic is trained jointly with the Actor using the PPO algorithm to minimize the value estimation error.

% \textcolor{red}{[TODO: No $\sigma$ function for critic in MetaBox]}

\subsection{Warm-starting and Context Memory}
Modern Evolutionary Computation algorithms typically maintain rich contextual information, such as dynamically adaptive parameters and runtime configurations, throughout the optimization process. For single-optimizer execution, such information is seamlessly updated throughout the optimization process. However, switching optimizers mid-process disrupts this continuity, as different optimizers rely on distinct contextual information and internal states. This gives rise to the warm-start problem—specifically, how to properly manage the context of different algorithms and perform context restoration for a smooth search. This issue has been explored in low-dimensional optimization, where a common solution is to maintain a shared context dictionary to store contextual memory.

In LSGO, CC inherently faces this challenge, as it optimizes distinct subproblems—a process that effectively entails switching between problem contexts and optimizers. However, this issue has largely been overlooked, primarily because precise decomposition in traditional LSGO problems typically yields significant performance gains that even overshadow these switching costs. In low-dimensional optimization, switching occurs between optimizers on the same problem. In traditional LSGO, the same optimizer switches between different subproblems. In H-LSGO, however, LH-CC must switch between different optimizers across different subproblems—a significantly more challenging scenario. To address this, leveraging established experience in low-dimensional optimization, we extend the shared context mechanism to the specific H-LSGO domain effectively as a proof-of-concept.
\begin{equation}
    \Gamma:\left\{\begin{array}{c}
    \begin{aligned}
\Gamma_{1}:&\left\{\begin{array}{c}
\Gamma_{1,1}:\left\{\text{ctx}_{1,1,1}, \cdots \right\} \\
\vdots \\
\Gamma_{1, k}:\left\{\text{ctx}_{1,k,1}, \cdots\right\} \\
\text {Common}_1:\left\{\text{ctx}_{1,1},  \cdots\right\}
\end{array}\right. \\
\vdots\\
\Gamma_{K}:&= \cdots
\end{aligned}
\end{array}\right.
\end{equation}
Here, $\text{ctx}_{i,j,*}$ denotes the specific context for the $j$-th optimizer on subproblem $f_i$, which is updated upon execution. The $\text{Common}_i$ dictionary stores shared contextual information accessible to all candidate optimizers for subproblem $f_i$, allowing updates from multiple sources. To restore an optimizer for subproblem $f_i$, the corresponding $\Gamma_{i,*}$ is retrieved, and the contexts within both $\Gamma_{i,*}$ and $\text{Common}_i$ are reloaded to facilitate the optimizer's warm-start. This restoration mechanism is generalizable to various algorithms, as it imposes no specific constraints on the context structure.
%     \Gamma:=\left\{\begin{array}{c}
% \Gamma_{1}:\left\{\operatorname{ctx}_{1,1}, \operatorname{ctx}_{1,2}, \cdots\right\} \\
% \vdots \\
% \Gamma_{M}:\left\{\operatorname{ctx}_{M, 1}, \operatorname{ctx}_{M, 2}, \cdots\right\} 
% \end{array}\right.

% The environment $\mathcal{E}$ maintains a problem set $\mathcal{P} = \{p_1, \dots, p_M\}$ and an algorithm pool $\mathcal{O} = \{o_1, \dots, o_L\}$. However, they alone are insufficient for handling the challenge of maintaining continuity across discrete decision steps in LH-CC. To address this, We introduce a \textit{Context Memory} mechanism to enable warm-starting.

% The interaction between the selected optimizer and the current solution is formalized as:
% \begin{equation}
% (x_t,\; c_t,\; C_t)
% = \text{Optimize}\!\left(
% o_{a_t},\;
% x_{t-1},\;
% C_{t-1}
% \right),
% \end{equation}
% where $x_t$ denotes the updated solution of the currently active subproblem at decision step $t$, $c_t$ is the corresponding objective value, and $C_t$ is the updated internal context of the optimizer. 
% The decision step $t$ is a \emph{global counter} shared across all subproblems, incrementing sequentially each time any subcomponent is optimized, rather than restarting for each subproblem.
% By passing $C_t$ to the next decision step $t+1$, LH-CC ensures that the internal state of adaptive algorithms remains consistent across steps, avoiding the "reset" penalty commonly observed in sequential optimization.

\section{Auto-H-LSGO Benchmark}
\label{sec:hlsgo}

\begin{figure}[h]
  \centering
  \includegraphics[width=\linewidth]{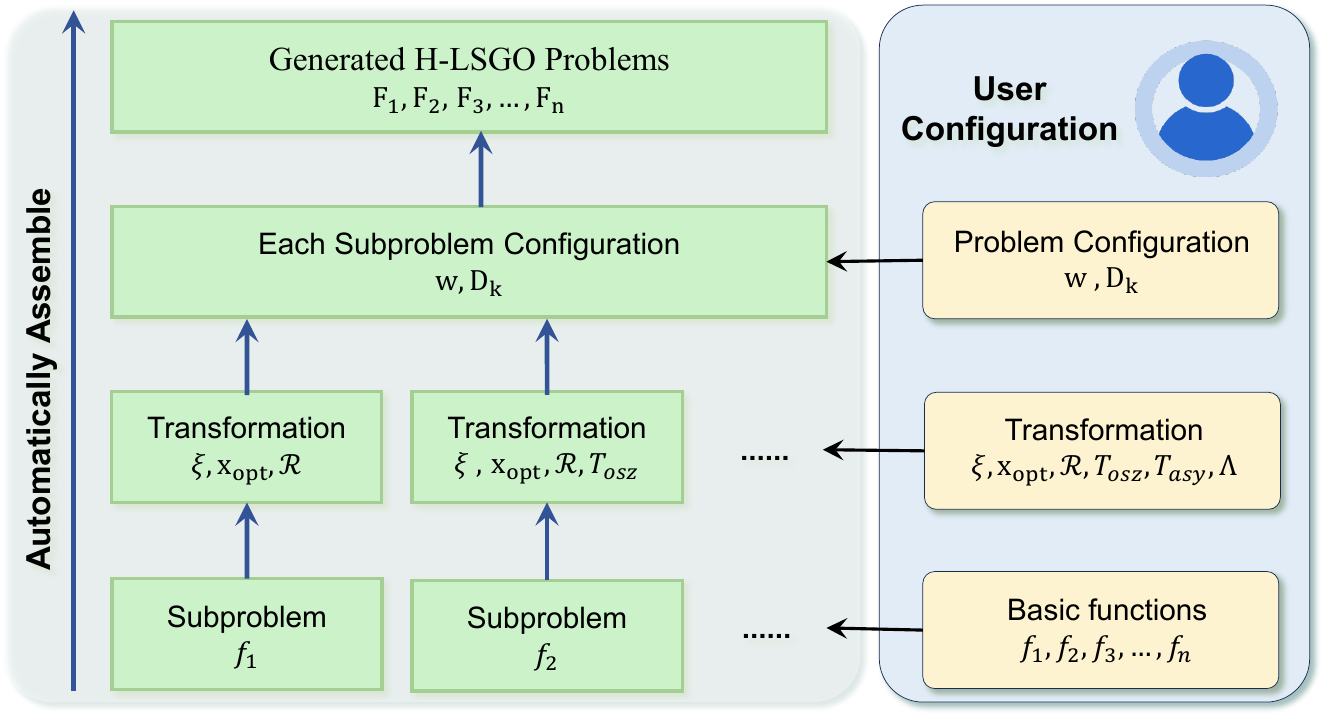}
  \caption{The generation pipeline for diverse problem instances within the Auto-H-LSGO.}
  \label{figure:hlsgo}
  \vspace{-16pt}
\end{figure}

As discussed in Section~\ref{sec:lsgo-cc}, while the CEC2013LSGO suite provides established benchmarks for high-dimensional optimization, its problem instances are primarily homogeneous in nature. 
Even though recent studies have attempted to introduce heterogeneous problems~\cite{xu2023large}, these instances are typically predefined, resulting in fixed structures and limited diversity, which hinders the agent's ability to learn across a broad spectrum of problems. 
The scarcity of problem instances represents an inherent challenge in the current LSGO domain. Given that H-LSGO further emphasizes subspace flexibility and diversity, this limitation can be addressed by designing a benchmark capable of automated design and assembly. By offering customizable configurations—such as overlap structures and subproblem types—this system can automatically generate a vast array of problem instances. 
Therefore, we propose our Auto-H-LSGO Benchmark, which enables the automated and systematic generation of diverse problem instances.
Figure~\ref{figure:hlsgo} illustrates the generation pipeline for diverse problem instances within the Auto-H-LSGO. 
To construct an optimization problem, users need only configure basic components according to their requirements. Auto-H-LSGO then automatically assembles these components to generate a series of problem instances ready for direct evaluation.
% In constructing an optimization problem, users can independently select the basic function for each subcomponent and apply optional transformation functions. Once the dimensionality and overlap ratios for the subproblems are specified, H-LSGO automatically assembles these components into a unified optimization problem. 

Specifically, we employ seven basic functions exhibiting distinct topological properties: Ackley, Elliptic, Rastrigin, Schwefel, Sphere, Katsuura, and Attractive Sector (with support for user-defined custom functions). 
For each subproblem, various transformation functions can be applied to manipulate the landscape characteristics. Specifically, a rotation matrix $\mathcal{R}$ is employed to enforce non-separability within the subproblem's dimensions. Meanwhile, the oscillated transformation $T_{osz}$ is utilized to introduce smooth local irregularities, the asymmetry transformation $T_{asy}$ is applied to break structural symmetry, and the diagonal scaling matrix $\Lambda$ is used to perform variable scaling for each dimension. Furthermore, Auto-H-LSGO allows for fine-grained control over the internal structure by permitting independent settings for dimensionality and overlap ratios for each subproblem. 
During the assembly phase, users need only select the subproblems; key parameters—including the variable permutation $\xi$, subproblem weights $w$, and the global optimum $\mathbf{x}_{opt}$—can then be either manually configured or automatically generated via a random seed.
This modular and automated design ensures that Auto-H-LSGO can produce highly complex, heterogeneous problem instances that closely mimic the challenges of real-world large-scale optimization. 
Detailed specifications of basic functions and the assembly process are provided in Appendix~\ref{app:benchmark_details}.

\section{Workflow}
\label{sec:workflow}

The workflow of LH-CC after training is completed is presented in Algorithm~\ref{alg:ccrldas}.
For each problem instance $F \in \Upsilon$ (Line 3), we decompose it into $K$ subproblems using $G$ (Line 4), enabling optimization on each subproblem. Then, the best solution $x^*$, best cost $c^*$, initial state $s_0$, context memory $\Gamma$, and decision step count $t$ are initialized (Line 5-6).
The core optimization loop will run sequentially through subproblems (Line 8), unless the FEs budget is exhausted or $c^*$ is less than 1E-20 (Line 7, 12). At each decision step $t$, when optimizing $k$-th subproblem, the actor network $\pi_\theta$ samples an action $a_t$ based on state $s_t$ (Line 9), which will be used as the index of chosen algorithm. This algorithm updates the $k$-th subproblem, yielding the improved global best solution $x^*$, next state $s_{t+1}$, reward $r_t$, and updated context $\Gamma$ (Line 10). With $x^*$ and $c^*$ gradually updated, the final best solution and best cost will be returned as the termination condition is met (Line 15).

The agent $\pi_\theta$ of LH-CC is trained using PPO~\cite{schulman2017proximal}.
At each decision step $t$, the target value for the critic $V_\phi$ is computed as an n-step bootstrapped discounted return using a backward-summation approach: $R_t = \sum_{i=0}^{n-1} \gamma^{i}r_{t+i} + \gamma^n V^{old}(s_{t+n})$. With value clipping $V_{clip}$ applied relative to the old value estimates $V^{old}$, the value loss is:
\begin{equation}
\mathcal{L}_V = \mathbb{E}[\max((V-R)^2, (V_{clip} - R)^2)].
\end{equation}
Then, for policy loss, we estimate $A_i$ with Generalized Advantage Estimation (GAE). Meanwhile, we store old log-probabilities $\log \pi_{\theta}^{old}(\cdot)$ and recompute log-probabilities under the current policy $\log \pi_{\theta}^{new}(\cdot)$ to form the importance sampling ratio
\begin{equation}
    \rho_t = \exp(\log \pi_{\theta}^{new}(a_t | \cdot) - \log \pi_{\theta}^{old}(a_t | \cdot)).
\end{equation}
The clipped surrogate policy loss is
\begin{equation}
    \mathcal{L}_{policy} = -\mathbb{E}[\min (\rho_t A_t, \text{clip} (\rho_t, 1 - \epsilon, 1 + \epsilon)A_t)] . 
\end{equation}
An entropy bonus is added to encourage exploration. The total objective is formulated as
\begin{equation}
    \mathcal{L} = \mathcal{L}_{policy} + \alpha \mathcal{L}_V - \beta \mathcal{H}(\pi_\theta).
\end{equation}
where $\alpha$ and $\beta$ weight the value function approximation error and entropy regularization respectively. We optimize the parameters using Adam, employing gradient norm clipping and an exponentially decayed learning rate. In each iteration, the framework collects $n_{\text{step}}$ transitions across parallel environments and performs $K_{\text{epoch}}$ optimization epochs over the same batch of data, where the number of epochs is dynamically scaled based on the remaining FEs.

\begin{algorithm}[t]
\caption{LH-CC}
\label{alg:ccrldas}
\begin{algorithmic}[1]
\STATE \textbf{Input:} Problem Set $\Upsilon$, Decomposition Method $\mathcal{G}$, Optimizer Pool $\mathcal{A}$, Pretrained Agent $\pi_\theta$.
\STATE \textbf{Output:} The Best Solution $x^*$, The Best Cost $c^*$
\FORALL{Problem Instance $F \in \Upsilon$}
    \STATE Decompose problem into $K$ subproblems using $\mathcal{G}$
    \STATE Initialize $x^*$, $c^*$, state $s_0$ and context memory $\Gamma$
    \STATE $t \leftarrow 0$
    \WHILE{Termination condition not met}
        \FOR{$k \leftarrow 1$ \textbf{to} $K$}
            \STATE $a_t \leftarrow \pi_\theta(s_t)$
            \STATE $\mathbf{x}^{\ast}, s_{t+1}, r_t, \Gamma \leftarrow \text{Step}(F, k, \mathbf{x}^{\ast}, \mathcal{A}_{a_t}, \Gamma)$
            % \STATE Store transition $(s_t, a_t, r_t, s_{t+1}, \log\pi_\theta(a_t|s_t))$
            \STATE $t \leftarrow t + 1$
            % \IF{Termination condition met}
            %     \STATE Break Optimization
            % \ENDIF
        \ENDFOR
    \ENDWHILE
\ENDFOR
\RETURN $x^*$, $c^*$

\end{algorithmic}
\end{algorithm}

\section{Experiment}
\label{sec:experiment}
Our experiments are designed to evaluate the effectiveness and robustness of LH-CC through the following key questions:

\begin{itemize}
  \item \textbf{RQ1:} How effective~(Section~\ref{sec:exp:perf}) and efficient~(Section~\ref{sec:exp:effi}) is LH-CC on H-LSGO? 
  \item \textbf{RQ2:} Can LH-CC function as a universally applicable framework independent of the optimizers~(Section~\ref{sec:exp:generalized})? 
  \item \textbf{RQ3:} Does the learned optimizer selection mechanism in LH-CC provide a meaningful advantage over fixed or random selection strategies (Section~\ref{sec:exp:ablation})? 
\end{itemize}

\subsection{Experimental Setup}

\subsubsection{Benchmark}
\label{sec:benchmark}
Based on Auto-H-LSGO, we constructed a comprehensive set of H-LSGO problem instances. To rigorously evaluate the performance of LH-CC, we categorized these instances into two distinct classes: one where subproblems share the same basic function but vary in dimensionality (e.g., mixing high and low dimensions), and another where subproblems differ in both basic function type and dimensionality. For the first category, we adopt the naming convention $\text{\{Basic Function\}}_{\{\text{Separability Degree}\}}$. Separability Degree is graded on a scale of 1 to 5, where 1 represents a fully separable problem and 5 denotes a fully non-separable one. For instance, $\text{Ackley}_{1}$ denotes an Ackley-based fully separate problem instance. For the second category, which is constructed by assembling the seven basic functions from Auto-H-LSGO in varying sequences, we use the convention $\text{He}_{\{\text{Separability Degree}\}}$. The total dimension for all problem instances is fixed at $3000$, comprising 14 subspaces with dimensions selected from five distinct levels: $\{25, 50, 100, 500, 1000\}$. Further details are provided in the Appendix ~\ref{sec:instance-detail}.

\subsubsection{Comparison Algorithms}
We categorize the comparison algorithms into four groups: a) \emph{LH-CC based algorithms}: We introduce an instantiation of LH-CC, termed LH-CC-X. In this variant, the optimizer pool $\mathcal{A}_\text{high}$ consists of SEP-CMAES~\cite{ros2008simple}, VkD-CMAES~\cite{akimoto2016projection}, and MMES~\cite{he2020mmes}, while $\mathcal{A}_\text{low}$ comprises CMAES~\cite{auger2012tutorial}, RMES~\cite{li2017simple}, and FCMAES~\cite{li2018fast}. For the decomposition method, we employ OEDG~\cite{tian2024enhanced}, recognized as one of the state-of-the-art decomposition techniques.
b) \emph{CC-based algorithms}: Both OEDG-CMAES and OEDG-MMES-CMAES employ OEDG as the decomposition strategy. However, they differ in their optimizer configuration: the former utilizes CMAES as the sole optimizer, whereas the latter adopts the LH-CC paradigm, using MMES as the high-dimensional optimizer and CMAES as the low-dimensional optimizer.
c) \emph{Non-Decomposition Algorithms} (NDAs): When contrasted with algorithms under the CC framework, high-dimensional optimizers are often collectively termed Non-Decomposition Algorithms (NDAs). This terminology underscores the fundamental distinction in their problem-solving paradigms. Here, we employ two representative algorithms: SEP-CMAES~\cite{ros2008simple} and MMES~\cite{he2020mmes}.

\begin{table}[htbp]
\centering
\caption{Comparing LH-CC with Comparison Algorithms on Auto-H-LSGO at 3E6 FEs}
\label{tab:main-exp1}
\renewcommand{\arraystretch}{1.2} 
\resizebox{\linewidth}{!}{
\begin{tabular}{c|c|cc|cc}
\hline
\multirow{2}{*}{Problem} & LH-CC-based & \multicolumn{2}{c|}{CC-based} & \multicolumn{2}{c}{NDAs} \\[0.5ex] \cline{2-6}
 & LH-CC-X & OEDG-CMAES & \makecell{OEDG-MMES\\- CMAES} & SEP-CMAES & MMES \\
\hline
    \multirow{2}{*}{$\text{Ackley}_1$} & \cellcolor[rgb]{0.400,0.400,0.400}\color{white}\textbf{2.16E+05} & \cellcolor[rgb]{1.000,1.000,1.000} 2.24E+05(+) & \cellcolor[rgb]{0.948,0.948,0.948} 2.20E+05(+) & \cellcolor[rgb]{0.967,0.967,0.967} 2.21E+05(+) & \cellcolor[rgb]{0.948,0.948,0.948} 2.20E+05(+) \\
     & \cellcolor[rgb]{0.400,0.400,0.400}\color{white}{$\pm$ 1.55E+03} & \cellcolor[rgb]{1.000,1.000,1.000}{$\pm$ 2.30E+03} & \cellcolor[rgb]{0.948,0.948,0.948}{$\pm$ 1.79E+03} & \cellcolor[rgb]{0.967,0.967,0.967}{$\pm$ 1.52E+02} & \cellcolor[rgb]{0.948,0.948,0.948}{$\pm$ 1.58E+03} \\
\hline
    \multirow{2}{*}{$\text{Ackley}_3$} & \cellcolor[rgb]{0.400,0.400,0.400}\color{white}\textbf{2.17E+05} & \cellcolor[rgb]{0.919,0.919,0.919} 2.19E+05(+) & \cellcolor[rgb]{0.919,0.919,0.919} 2.19E+05(+) & \cellcolor[rgb]{1.000,1.000,1.000} 2.23E+05(+) & \cellcolor[rgb]{0.948,0.948,0.948} 2.20E+05(+) \\
     & \cellcolor[rgb]{0.400,0.400,0.400}\color{white}{$\pm$ 4.63E+02} & \cellcolor[rgb]{0.919,0.919,0.919}{$\pm$ 1.68E+03} & \cellcolor[rgb]{0.919,0.919,0.919}{$\pm$ 1.22E+03} & \cellcolor[rgb]{1.000,1.000,1.000}{$\pm$ 1.93E+02} & \cellcolor[rgb]{0.948,0.948,0.948}{$\pm$ 9.40E+02} \\
\hline
    \multirow{2}{*}{$\text{Ackley}_5$} & \cellcolor[rgb]{0.400,0.400,0.400}\color{white}\textbf{2.19E+05} & \cellcolor[rgb]{1.000,1.000,1.000} 2.26E+05(+) & \cellcolor[rgb]{0.959,0.959,0.959} 2.23E+05(+) & \cellcolor[rgb]{0.977,0.977,0.977} 2.24E+05(+) & \cellcolor[rgb]{0.910,0.910,0.910} 2.21E+05(+) \\
     & \cellcolor[rgb]{0.400,0.400,0.400}\color{white}{$\pm$ 5.06E+02} & \cellcolor[rgb]{1.000,1.000,1.000}{$\pm$ 4.90E+02} & \cellcolor[rgb]{0.959,0.959,0.959}{$\pm$ 4.21E+02} & \cellcolor[rgb]{0.977,0.977,0.977}{$\pm$ 5.80E+01} & \cellcolor[rgb]{0.910,0.910,0.910}{$\pm$ 2.59E+03} \\
\hline
    \multirow{2}{*}{$\text{Attractive}_1$} & \cellcolor[rgb]{0.400,0.400,0.400}\color{white}\textbf{1.02E-04} & \cellcolor[rgb]{1.000,1.000,1.000} 4.44E+32(+) & \cellcolor[rgb]{0.889,0.889,0.889} 4.59E+02(+) & \cellcolor[rgb]{0.996,0.996,0.996} 1.01E+29(+) & \cellcolor[rgb]{0.896,0.896,0.896} 9.99E+03(+) \\
     & \cellcolor[rgb]{0.400,0.400,0.400}\color{white}{$\pm$ 1.21E-04} & \cellcolor[rgb]{1.000,1.000,1.000}{$\pm$ 6.42E+31} & \cellcolor[rgb]{0.889,0.889,0.889}{$\pm$ 1.25E+02} & \cellcolor[rgb]{0.996,0.996,0.996}{$\pm$ 3.28E+28} & \cellcolor[rgb]{0.896,0.896,0.896}{$\pm$ 4.92E+02} \\
\hline
    \multirow{2}{*}{$\text{Attractive}_2$} & \cellcolor[rgb]{0.400,0.400,0.400}\color{white}\textbf{1.31E-04} & \cellcolor[rgb]{0.995,0.995,0.995} 9.12E+25(+) & \cellcolor[rgb]{0.881,0.881,0.881} 1.02E+01(+) & \cellcolor[rgb]{1.000,1.000,1.000} 3.52E+29(+) & \cellcolor[rgb]{0.900,0.900,0.900} 9.69E+03(+) \\
     & \cellcolor[rgb]{0.400,0.400,0.400}\color{white}{$\pm$ 2.50E-04} & \cellcolor[rgb]{0.995,0.995,0.995}{$\pm$ 1.65E+25} & \cellcolor[rgb]{0.881,0.881,0.881}{$\pm$ 8.50E+00} & \cellcolor[rgb]{1.000,1.000,1.000}{$\pm$ 9.99E+28} & \cellcolor[rgb]{0.900,0.900,0.900}{$\pm$ 4.44E+02} \\
\hline
    \multirow{2}{*}{$\text{Attractive}_5$} & \cellcolor[rgb]{0.400,0.400,0.400}\color{white}\textbf{2.58E+01} & \cellcolor[rgb]{0.997,0.997,0.997} 1.08E+27(+) & \cellcolor[rgb]{0.875,0.875,0.875} 3.37E+04(+) & \cellcolor[rgb]{1.000,1.000,1.000} 2.07E+29(+) & \cellcolor[rgb]{0.868,0.868,0.868} 4.03E+03(+) \\
     & \cellcolor[rgb]{0.400,0.400,0.400}\color{white}{$\pm$ 1.57E+00} & \cellcolor[rgb]{0.997,0.997,0.997}{$\pm$ 1.44E+26} & \cellcolor[rgb]{0.875,0.875,0.875}{$\pm$ 1.13E+04} & \cellcolor[rgb]{1.000,1.000,1.000}{$\pm$ 8.01E+28} & \cellcolor[rgb]{0.868,0.868,0.868}{$\pm$ 3.57E+02} \\
\hline
    \multirow{2}{*}{$\text{Katsuura}_1$} & \cellcolor[rgb]{0.851,0.851,0.851} 2.18E+06 & \cellcolor[rgb]{1.000,1.000,1.000} 4.44E+32(+) & \cellcolor[rgb]{0.852,0.852,0.852} 2.54E+06(+) & \cellcolor[rgb]{0.400,0.400,0.400}\color{white}\textbf{1.48E+06(-)} & \cellcolor[rgb]{0.860,0.860,0.860} 2.02E+07(+) \\
     & \cellcolor[rgb]{0.851,0.851,0.851}{$\pm$ 1.04E+05} & \cellcolor[rgb]{1.000,1.000,1.000}{$\pm$ 5.30E+31} & \cellcolor[rgb]{0.852,0.852,0.852}{$\pm$ 1.60E+05} & \cellcolor[rgb]{0.400,0.400,0.400}\color{white}{$\pm$ 1.45E+05} & \cellcolor[rgb]{0.860,0.860,0.860}{$\pm$ 9.44E+05} \\
\hline
    \multirow{2}{*}{$\text{Katsuura}_4$} & \cellcolor[rgb]{0.400,0.400,0.400}\color{white}\textbf{8.88E+06} & \cellcolor[rgb]{1.000,1.000,1.000} 9.12E+25(+) & \cellcolor[rgb]{0.852,0.852,0.852} 1.32E+07(+) & \cellcolor[rgb]{0.853,0.853,0.853} 1.49E+07(+) & \cellcolor[rgb]{0.856,0.856,0.856} 2.66E+07(+) \\
     & \cellcolor[rgb]{0.400,0.400,0.400}\color{white}{$\pm$ 1.18E+06} & \cellcolor[rgb]{1.000,1.000,1.000}{$\pm$ 1.28E+25} & \cellcolor[rgb]{0.852,0.852,0.852}{$\pm$ 2.45E+06} & \cellcolor[rgb]{0.853,0.853,0.853}{$\pm$ 2.89E+06} & \cellcolor[rgb]{0.856,0.856,0.856}{$\pm$ 2.17E+06} \\
\hline
    \multirow{2}{*}{$\text{Katsuura}_5$} & \cellcolor[rgb]{0.400,0.400,0.400}\color{white}\textbf{1.81E+07} & \cellcolor[rgb]{1.000,1.000,1.000} 1.08E+27(+) & \cellcolor[rgb]{0.853,0.853,0.853} 3.44E+07(+) & \cellcolor[rgb]{0.857,0.857,0.857} 6.77E+07(+) & \cellcolor[rgb]{0.855,0.855,0.855} 5.38E+07(+) \\
     & \cellcolor[rgb]{0.400,0.400,0.400}\color{white}{$\pm$ 5.62E+05} & \cellcolor[rgb]{1.000,1.000,1.000}{$\pm$ 9.83E+25} & \cellcolor[rgb]{0.853,0.853,0.853}{$\pm$ 1.40E+06} & \cellcolor[rgb]{0.857,0.857,0.857}{$\pm$ 1.65E+05} & \cellcolor[rgb]{0.855,0.855,0.855}{$\pm$ 1.24E+06} \\
\hline
\hline
    \multirow{2}{*}{$\text{He}_1$} & \cellcolor[rgb]{0.400,0.400,0.400}\color{white}\textbf{8.21E+06} & \cellcolor[rgb]{0.977,0.977,0.977} 1.73E+18(+) & \cellcolor[rgb]{0.941,0.941,0.941} 1.81E+14(+) & \cellcolor[rgb]{1.000,1.000,1.000} 6.57E+22(+) & \cellcolor[rgb]{0.886,0.886,0.886} 3.85E+09(+) \\
     & \cellcolor[rgb]{0.400,0.400,0.400}\color{white}{$\pm$ 1.14E+06} & \cellcolor[rgb]{0.977,0.977,0.977}{$\pm$ 4.14E+16} & \cellcolor[rgb]{0.941,0.941,0.941}{$\pm$ 3.26E+13} & \cellcolor[rgb]{1.000,1.000,1.000}{$\pm$ 4.78E+21} & \cellcolor[rgb]{0.886,0.886,0.886}{$\pm$ 6.17E+07} \\
\hline
    \multirow{2}{*}{$\text{He}_2$} & \cellcolor[rgb]{0.400,0.400,0.400}\color{white}\textbf{2.35E+07} & \cellcolor[rgb]{0.977,0.977,0.977} 8.07E+17(+) & \cellcolor[rgb]{0.887,0.887,0.887} 8.19E+09(+) & \cellcolor[rgb]{1.000,1.000,1.000} 1.14E+22(+) & \cellcolor[rgb]{0.883,0.883,0.883} 4.41E+09(+) \\
     & \cellcolor[rgb]{0.400,0.400,0.400}\color{white}{$\pm$ 2.60E+07} & \cellcolor[rgb]{0.977,0.977,0.977}{$\pm$ 9.23E+17} & \cellcolor[rgb]{0.887,0.887,0.887}{$\pm$ 7.56E+09} & \cellcolor[rgb]{1.000,1.000,1.000}{$\pm$ 4.37E+21} & \cellcolor[rgb]{0.883,0.883,0.883}{$\pm$ 5.19E+08} \\
\hline
    \multirow{2}{*}{$\text{He}_3$} & \cellcolor[rgb]{0.400,0.400,0.400}\color{white}\textbf{2.37E+08} & \cellcolor[rgb]{0.988,0.988,0.988} 3.65E+19(+) & \cellcolor[rgb]{0.920,0.920,0.920} 1.06E+13(+) & \cellcolor[rgb]{1.000,1.000,1.000} 1.37E+22(+) & \cellcolor[rgb]{0.870,0.870,0.870} 4.10E+09(+) \\
     & \cellcolor[rgb]{0.400,0.400,0.400}\color{white}{$\pm$ 7.97E+07} & \cellcolor[rgb]{0.988,0.988,0.988}{$\pm$ 7.30E+19} & \cellcolor[rgb]{0.920,0.920,0.920}{$\pm$ 2.14E+12} & \cellcolor[rgb]{1.000,1.000,1.000}{$\pm$ 1.27E+22} & \cellcolor[rgb]{0.870,0.870,0.870}{$\pm$ 2.66E+08} \\
\hline
    \multirow{2}{*}{$\text{He}_4$} & \cellcolor[rgb]{0.400,0.400,0.400}\color{white}\textbf{1.36E+08} & \cellcolor[rgb]{0.964,0.964,0.964} 1.22E+17(+) & \cellcolor[rgb]{0.910,0.910,0.910} 2.09E+12(+) & \cellcolor[rgb]{1.000,1.000,1.000} 6.88E+22(+) & \cellcolor[rgb]{0.873,0.873,0.873} 4.13E+09(+) \\
     & \cellcolor[rgb]{0.400,0.400,0.400}\color{white}{$\pm$ 5.03E+07} & \cellcolor[rgb]{0.964,0.964,0.964}{$\pm$ 5.95E+16} & \cellcolor[rgb]{0.910,0.910,0.910}{$\pm$ 1.40E+11} & \cellcolor[rgb]{1.000,1.000,1.000}{$\pm$ 4.03E+22} & \cellcolor[rgb]{0.873,0.873,0.873}{$\pm$ 2.78E+08} \\
\hline
    \multirow{2}{*}{$\text{He}_5$} & \cellcolor[rgb]{0.400,0.400,0.400}\color{white}\textbf{2.90E+08} & \cellcolor[rgb]{0.956,0.956,0.956} 1.52E+16(+) & \cellcolor[rgb]{0.865,0.865,0.865} 2.62E+09(+) & \cellcolor[rgb]{1.000,1.000,1.000} 2.56E+22(+) & \cellcolor[rgb]{0.867,0.867,0.867} 3.20E+09(+) \\
     & \cellcolor[rgb]{0.400,0.400,0.400}\color{white}{$\pm$ 6.82E+07} & \cellcolor[rgb]{0.956,0.956,0.956}{$\pm$ 9.94E+15} & \cellcolor[rgb]{0.865,0.865,0.865}{$\pm$ 1.84E+08} & \cellcolor[rgb]{1.000,1.000,1.000}{$\pm$ 1.71E+22} & \cellcolor[rgb]{0.867,0.867,0.867}{$\pm$ 3.44E+08} \\
\hline
\multicolumn{1}{c|}{+/$\approx$/-} & N/A & 14/0/0 & 14/0/0 & 13/0/1 & 14/0/0  \\
\hline
 $\Delta \sum \log_{10}$ & N/A & -206.95 & -34.88 & -167.99 & -28.70 \\
\hline
\end{tabular}
}
\end{table}

\subsubsection{Basic Settings of Algorithms}
LH-CC-X is trained for 30 epochs, with each epoch allocated $\text{MaxFEs}=1\text{E+06}$. We opted for $1\text{E+06}$ rather than the standard $3\text{E+06}$ commonly used in LSGO benchmarks, as the latter incurs prohibitive computational costs during training. Consequently, adopting the reduced budget serves a dual purpose: it mitigates computational overhead and allows us to verify the generalization capability of LH-CC across different FE horizons. Furthermore, we performed training on dozens of heterogeneous problems constructed using only Ackley, Elliptic, Schwefel, and Rastrigin as basic functions, with separability degree restricted to 1, 2, and 3. Compared to the test set, this configuration offers limited diversity in both basic functions and separability levels, thereby facilitating the verification of LH-CC-X's cross-problem generalization capability. Each step in LH-CC-X receives 2500 FEs and decomposition results are precomputed to accelerate training. The training is conducted with other settings including $n_{\text{step}} = 10$, $K_{\text{epoch}}$ $= 12$, $\gamma = 0.99$, and learning rate $=$ 1E-05. In addition to optimizer-specific parameters, $\Gamma$ contains a $Common$ component that stores the global-best solution $x^*$, the population mean $\bar{x}$, and the step size $\sigma$. All comparison algorithms are configured according to their original papers, with optimizers implemented using open-source libraries such as pypop7 \cite{duan2024pypop7,lange2023evosax}.

\begin{table}[htbp]
\centering
\caption{Comparison of Optimization Efficiency between LH-CC and Comparative Algorithms}
\label{tab:main-exp2}
\renewcommand{\arraystretch}{1.2} 
\resizebox{\linewidth}{!}{
\begin{tabular}{c|c|cc|cc}
\hline
\multirow{2}{*}{Problem Type} & LH-CC-based & \multicolumn{2}{c|}{CC-based} & \multicolumn{2}{c}{NDAs} \\[0.5ex] \cline{2-6}
 & LH-CC-X & OEDG-CMAES & \makecell{OEDG-MMES\\- CMAES} & SEP-CMAES & MMES \\
    \hline
1 & 1095.51s & 13191.61s & 957.43s & 796.19s & 580.14s \\
\hline \hline 
2 & 1138.62s & 14575.32s & 1051.2s & 787.46s & 586.47s \\
\hline
\end{tabular}
}
\end{table}

\subsection{Comparative Analysis}
\label{sec:exp:main-exp}

\subsubsection{Optimization Performance Comparison \textbf{(RQ1)}}
\label{sec:exp:perf}
Table~\ref{tab:main-exp1} reports the performance on multiple problem instances instantiated via the Auto-H-LSGO benchmark with $\text{MaxFEs} =$ 3E+06. The symbols “+”, “$-$”, and “$\approx$” denote significantly superior, inferior, and comparable performance (Wilcoxon rank-sum test, $\alpha=0.05$), respectively, with aggregated counts (win/tie/loss) summarized in the penultimate row. The final row presents the total difference in orders of magnitude between the objective values of the comparative algorithms and LH-CC-X (calculated as the sum of log-scale differences across all instances $\Delta \sum \log_{10}$).

% The performance results are summarized in Table~\ref{tab:main-exp}, where the values denote the final global best cost $c^*$ achieved within the maximum evaluation budget. For each problem instance, the best performance among all tested algorithms is highlighted in \textbf{bold}. Also, the total computational time is also summarized, providing a crucial metric for evaluating the practical efficiency of the algorithms, especially in large-scale scenarios.
% To provide a clear quantitative assessment of the relative performance, the final row of Table~\ref{tab:main-exp} reports the overall performance of LH-CC against each baseline. This summary is formatted as ``W/T/L $|$ $\Delta \sum \log_{10}$'', representing the count of wins, ties, and losses for LH-CC, followed by the total difference in the order of magnitude of the objective values (calculated as the sum of log-scale differences across all instances). For instance, the entry ``17/1/0 $|$ -33.54'' for DG2-CMAES indicates that LH-CC achieves 17 wins and 1 ties against the MMES, while our cumulative log-scale cost is 33.54 lower, demonstrating a significant overall improvement in optimization quality.

Based on the results presented in Table~\ref{tab:main-exp1}, we observe the following key findings:
\begin{itemize}
    \item \textbf{Superior Optimization Performance}: It is evident that OEDG-CMAES, which exhibits remarkable optimization performance on homogeneous LSGO tasks, proves nearly ineffective on H-LSGO problems. In contrast, LH-CC-X significantly outperforms the majority of baselines across diverse problem instances, demonstrating its capability to navigate complex, heterogeneous LSGO problems.
    
    \item \textbf{Generalization across Optimization Horizons and Problems}: Despite being trained with $\text{MaxFEs}=$ 1E+06, LH-CC-X demonstrates significant superiority when tested at $\text{MaxFEs}=$ 3E+06 on instances that diverge from the training set in both basic function types and degrees of separability. This performance underscores its generalizability.
    
    \item \textbf{The Essential Role of Problem Decomposition}: NDAs, such as MMES, exhibit promising results on the 1000 dimensional problems of the CEC2013LSGO. However, their performance deteriorates significantly on the 3000 dimensional instances presented here. This decline is attributed to both the scalability bottleneck caused by increased dimensionality and the intrinsic complexity of the problems. NDAs fail to exploit structural features to mitigate difficulty, thereby underscoring the importance of decomposition.

    \item \textbf{Significance of Dynamic Optimizer Selection}: OEDG-MMES-CMAES, which is similarly equipped with distinct high- and low-dimensional optimizers, lags significantly behind LH-CC-X in performance. This discrepancy highlights the necessity of dynamic selection, a topic we will analyze in greater depth in Section \ref{sec:exp:ablation}.
\end{itemize}

\subsubsection{Optimization Efficiency Comparison \textbf{(RQ1)}}
\label{sec:exp:effi}

In Section \ref{sec:benchmark}, we categorized the problem instances into two classes: Type 1 (dimension extension only) and Type 2 (extension of both dimension and problem type). Concurrently with the experiments for Table \ref{tab:main-exp1}, we recorded the runtime of each algorithm. The results were averaged based on problem type and are summarized in Table \ref{tab:main-exp2}.

Table \ref{tab:main-exp2} provides three key insights. a) A comparison of LH-CC-based and CC-based algorithms indicates that the runtime penalty is largely driven by the mismatch between optimizer capabilities and subproblem dimensions. b) Comparing LH-CC-X with OEDG-MMES-CMAES, we observe that despite the introduction of the dynamic selection mechanism, the computational time does not increase significantly. c) NDAs demonstrate certain advantages in computational efficiency as they bypass the decomposition entirely.

\begin{figure}[t]
  \centering
  \includegraphics[width=\linewidth]{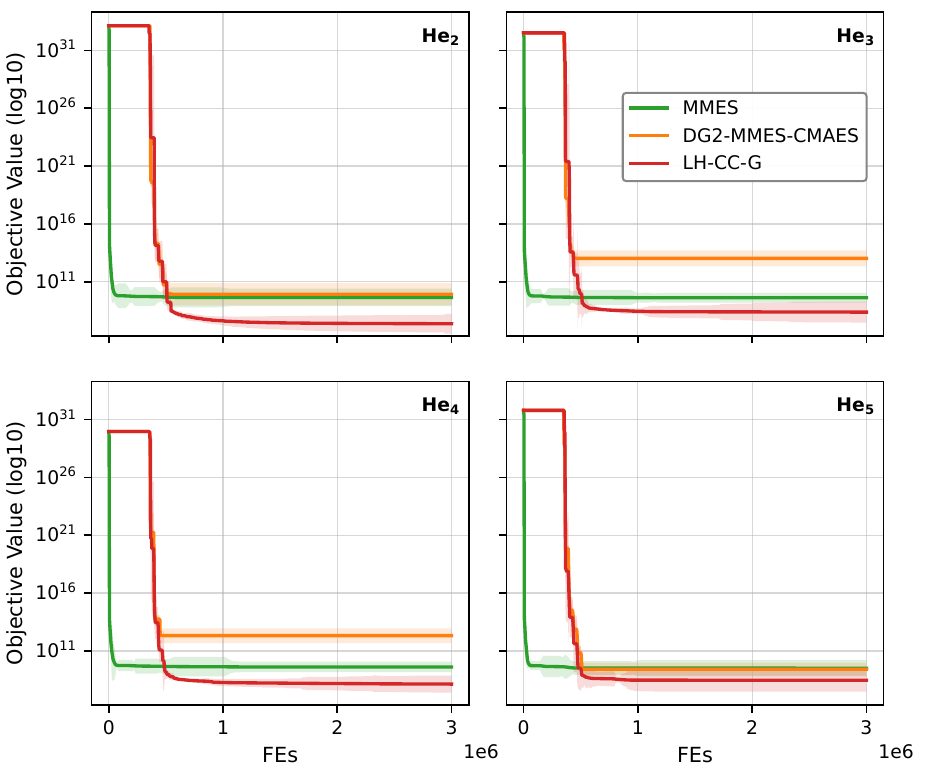}
  \caption{Comparative Optimization Performance of LH-CC-G on Representative H-LSGO Problems}
  \label{figure:generalized}
\end{figure}
\vspace{-6pt}

% \subsubsection{Generalization across Problems}
% The robust generalization across problem landscapes is inherently demonstrated by our experimental design. As detailed in the training configuration, the RL agent was trained exclusively on a small subset of homogeneous functions with limited overlap ratios. However, the performance results in Section~\ref{sec:exp:training} show that the learned policy maintains high efficacy when applied to the test set, which contains complex heterogeneous compositions. This indicates that LH-CC has learned generic features of subproblem difficulty and inter-component dependencies rather than over-fitting to specific function landscapes.

\subsubsection{Generalization across Optimizers \textbf{(RQ2)}}
\label{sec:exp:generalized}

Although the optimizers employed in LH-CC-X are predominantly from the ES family \cite{emmerich2025evolution}, LH-CC is designed as an algorithm-agnostic, general-purpose framework. To validate this generalizability, we directly replace the constituent optimizers within LH-CC-X with algorithms from the Differential Evolution (DE) \cite{pant2020differential} and Particle Swarm Optimization (PSO) families \cite{wang2018particle}. For clarity, we denote this variant as LH-CC-G. Specifically, the optimizer pools are reconfigured as follows: $\mathcal{A}_{\text{low}}$: comprising FCMAES, GLGA~\cite{garcia2008global}, and CLPSO~\cite{liang2006comprehensive}; $\mathcal{A}_{\text{high}}$ : comprising VkD-CMAES, IPSO~\cite{li2006improved}, and NL-SHADE-RSP~\cite{stanovov2021nl}. The comparative performance results of LH-CC-G on representative problem instances are illustrated in Figure~\ref{figure:generalized}. Experimental analysis reveals that even with fundamentally different underlying optimizers, LH-CC maintains significant performance advantages over static baseline configurations. This demonstrates that the agent successfully acquires a high-level coordination strategy capable of leveraging the complementary strengths of diverse optimizers.

% While the backbone optimizers used in our primary experiments belong to the ES family, the LH-CC framework is designed to be algorithm-agnostic. To verify this, we introduce algorithms from the Differential Evolution (DE) and Particle Swarm Optimization (PSO) families. Specifically, we re-configure the optimizer pools as follows:
% $\mathcal{A}_{\text{low}}$: Includes 
% % JADE~\cite{zhang2009jade}, 
% FCMAES, GLGA~\cite{garcia2008global}, and CLPSO~\cite{liang2006comprehensive};
% $\mathcal{A}_{\text{high}}$: Includes VkD-CMAES, IPSO~\cite{li2006improved}, and NL\_SHADE\_RSP~\cite{stanovov2021nl}.

% The comparative results of this configuration are presented in Figure~\ref{figure:ablation}. Experimental analysis reveals that even when the underlying search mechanisms are fundamentally changed, LH-CC still achieves significant performance improvements over static baseline configurations. This demonstrates that the agent has successfully learned a high-level coordination logic that can effectively leverage the complementary strengths of diverse optimizers.

\subsection{Ablation Study \textbf{(RQ3)}}
\label{sec:exp:ablation}

\begin{figure}[t]
  \centering
  \includegraphics[width=\linewidth]{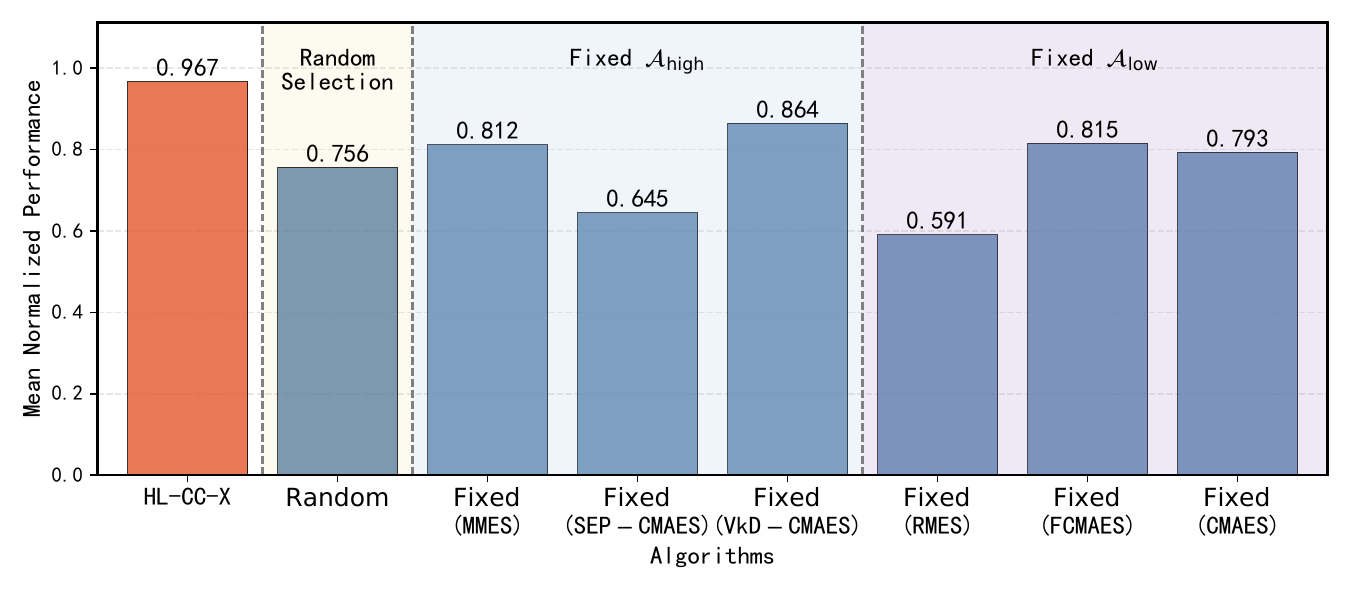}
  \caption{Ablation Study on the Action Space}
  \label{figure:ablation}
  \vspace{-10pt}
\end{figure}

To validate the efficacy of the dynamic optimizer selection mechanism in LH-CC, we conducted an ablation study comparing LH-CC-X against two baseline selection strategies: a) Random Selection: At each decision step, an optimizer is sampled uniformly at random from the candidate pool $\mathcal{A}$ for each subproblem. This baseline serves to determine whether performance gains are attributed solely to the diversity inherent in the optimizer pool. b) Fixed Selection: A single backbone optimizer is designated from the respective pool ($\mathcal{A}_\text{low}$ or $\mathcal{A}_\text{high}$) and employed exclusively throughout the entire optimization process. 
Here, instead of retraining, we directly enforced fixed actions on the pre-trained LH-CC-X.

The performance comparison is summarized in Figure~\ref{figure:ablation}. The experimental results yield several key insights. First, LH-CC significantly outperforms the Random Selection strategy, confirming that the meta-agent has successfully acquired meaningful state-action mappings. Second, LH-CC achieves superior results compared to the best-performing Fixed Selection baseline, a finding that aligns with the comparative analysis against OEDG-MMES-CMAES presented in Section \ref{sec:exp:perf}. This indicates that no single optimizer suffices across all search stages and subproblem types within heterogeneous landscapes. By dynamically switching between optimizers, LH-CC effectively harnesses the complementary search behaviors of diverse optimizers, achieving a synergistic effect that transcends the limitations of any individual backbone. This reveals that for H-LSGO problems, it is crucial not only to focus on the alignment between the optimizer and subproblem dimensionality but also to dynamically switch optimizers to match the distinct optimization characteristics of heterogeneous subproblems.

% To verify the effectiveness of the learned dynamic operator selection mechanism in LH-CC, we conduct an ablation study by comparing the RL-based meta-policy with two baseline selection strategies:
% a) Random Selection: At each decision step, an optimizer is sampled uniformly at random from the pool $\mathcal{A}$ for each subproblems. This baseline tests whether the performance gain is merely due to the diversity of the optimizer pool.
% b) Fixed Selection: A single specific backbone optimizer is consistently selected from its corresponding optimizer pool ($\mathcal{A}_\text{low}$ or $\mathcal{A}_\text{high}$) and applied throughout the entire optimization process.

% The performance comparison is summarized in Figure~\ref{fig:ablation}. The experimental results lead to several key conclusions:
% First, LH-CC significantly outperforms the \textit{Random Selection} strategy, demonstrating that the RL agent has indeed captured meaningful state-action mappings. Second, LH-CC achieves superior or competitive results compared to the best \textit{Fixed Selection} baseline on most instances. This indicates that no single optimizer can dominate across all search stages and subproblem types in heterogeneous landscapes. By adaptively switching between experts, LH-CC successfully harnesses the complementary search behaviors of different algorithms, achieving a "synergistic effect" that surpasses the limits of any individual backbone.

\section{Conclusion and Future Work}
\label{sec:conclusion}
In this paper, we propose LH-CC, a learning-based cooperative co-evolution framework designed to address the emerging challenges of heterogeneity in LSGO. By formulating the dynamic selection of optimizers as a Markov Decision Process, LH-CC enables automated coordination of diverse optimizers for different subproblems without relying on manual heuristic rules. Extensive evaluations on the proposed benchmark demonstrate that LH-CC consistently outperforms representative baselines on complex H-LSGO problems with dimensions up to 3000 and varying coupling relationships. Furthermore, our framework exhibits robust generalization capabilities across unseen problem landscapes and alternative optimizer families, enabling a 'train-once, apply-widely' paradigm that reduces the computational overhead for H-LSGO problems. Finally, ablation studies confirm that dynamic optimizer selection is a pivotal strategy for solving H-LSGO problems.

Future work can build upon the LH-CC framework in two main directions. On one hand, mechanisms such as subspace contribution-based resource allocation could be introduced to achieve more efficient distribution of computational resources. On the other hand, advanced architectures like auto-regressive networks could be employed, enabling the agent to autonomously determine when to deploy high- or low-dimensional optimizers, thereby obviating the need for manual categorization. As a pioneering attempt to address the H-LSGO problem, LH-CC has successfully validated its effectiveness as a substantial extension of the existing CC paradigm.
% In this paper, we propose LH-CC, a learning-based cooperative coevolution framework designed to address the challenges of  heterogeneity in large-scale global optimization. By formulating the dynamic selection of optimizers as a Markov Decision Process, LH-CC enables a fine-grained, automated coordination of diverse search strategies for different subcomponents without relying on manual heuristic rules. Extensive evaluations on the proposed benchmark demonstrate that LH-CC consistently outperforms representative baselines, particularly in complex scenarios with varying separability and overlapping structures. Furthermore, our framework exhibits robust generalization capabilities across unseen problem landscapes and alternative optimizer families, allowing for a "train-once, apply-widely" paradigm that significantly enhances computational efficiency. Finally, ablation studies confirm the superiority of our RL-based meta-policy over fixed or random selection strategies, providing new insights into the synergistic potential of hybridizing multiple optimization paradigms in a spatially-aware manner.

\begin{acks}
This work was supported in part by the Guangdong Provincial Natural Science Foundation for Outstanding Youth Team Project (Grant No. 2024B1515040010), in part by Guangzhou Science and Technology Elite Talent Leading Program for Basic and Applied Basic Research (Grant No. SL2024A04J01361), in part by the Fundamental Research Funds for the Central Universities (Grant No. 2025ZYGXZR027).
\end{acks}

\bibliographystyle{ACM-Reference-Format}
\bibliography{ref}


\appendix

\include{appendix}

\end{document}

%% file: appendix.tex
\begin{center}
    \textbf{\Large Appendix}
\end{center}
\section{Definition of Features}
\label{sec:feature}

\subsection{Problem Features}

\subsubsection{Dimension Indicator}

To indicate the relative size of subproblem, we introduce a feature to reflect the dimension. Let $D_k$ denote the dimension of the $k$-th subproblem $\mathbf{x}^{(k)}$, namely $D_k = \dim\big(\mathbf{x}^{(k)}\big)$. Then,
\begin{equation}
\mathcal{S}_{\text{problem}, 1}
=
\left(\frac{D_k}{500}\right)^{0.4}.
\label{feature:size}
\end{equation}
This feature is also used in action part, namely section \ref{sec:method:action}.

\subsubsection{Separation Indicator}

We include a binary feature indicating that whether the problem is separable, and it is derived from the Design Structure Matrix (DSM). If $\sum_{j=1}^{D} \Theta[i,j] = 1$, then \(\Theta[i,i] = 1\) and all other elements in the \(i\)-th row are zero, which indicates that variable \(i\) is completely independent.

The separation indicator is formally defined as follows:
\begin{equation}
\mathcal{S}_{\text{problem}, 2} = \mathcal{I}(\left| \left\{ i \mid \sum_{j=1}^{D} \Theta[i,j] = 1 \right\} \right| = D).
\label{feature:seperable}
\end{equation}

Here, \(\mathcal{I}\) is an indicator function that returns 1 if the condition inside the parentheses is true, and 0 otherwise. The condition checks if the variables are fully separable.

\subsubsection{Overlap Indicator}

To quantify the coupling intensity of subproblems, we define an overlap indicator based on the shared variables. Let \( \Omega_i \) and \( \Omega_j \) denote the variable sets of subproblem \( i \) and \( j \), respectively. The overlap feature is calculated as follows:

\begin{equation}
\mathcal{S}_{\text{problem}, 3} = \frac{\left| \bigcup_{i=1}^{|S|} \bigcup_{j \neq i}^{|S|} \left( \Omega_i \cap \Omega_j \right) \right|}{D}.
\label{feature:overlap}
\end{equation}

Here, \( \Omega_i \cap \Omega_j \) represents the intersection of variable sets between subproblems \( i \) and \( j \), and \( \bigcup \) denotes the union of all such intersections across all pairs of subproblems. The numerator counts the total number of shared variables across all subproblem pairs, while the denominator normalizes this count by the total number of variables \( D \), providing a measure of the average overlap.

% ============================================================
\subsection{Population Features}

\subsubsection{Dispersion and Dispersion Ratio}
To measure population diversity, we compute the average pairwise distance
\begin{equation}
\mathcal{S}_{\text{pop}, 1} = d = \frac{1}{N(N-1)} \sum_{i\neq j} \|x_i - x_j\|.
\label{feature:dispersion-def}
\end{equation}
Let $d_{\mathrm{top}}$ denote the average pairwise distance among the top $10\%$
best individuals. We define:
\begin{equation}
\mathcal{S}_{\text{pop}, 2} = d_{\mathrm{top}} - d.
\label{feature:dispersion}
\end{equation}
This reflects the landscape ``funnelity'':
a single-funnel landscape yields a small difference, while multi-funnel landscapes
produce larger values.

\subsubsection{Average Neutral Ratio (ANR)}
This feature is used to measure landscape ruggedness and evolvability.  
Let $C=\{c_{(i)}\}_{i=1}^{N}$ be the current population costs, where $c_{(i)}$ stands for the $i$-th individual in the population.
To probe local neutrality, we randomly pick one
optimizer $\mathcal{A}_l$ from the pool and let it advance the whole population by
\textit{one} optimization step.
We sample $S$ times, then we get $\{C'_s\}_{s=1}^S$ , where 
$C'_s=\{c^s_{(i)}\}_{i=1}^N$.
The average neutral ratio is:
\begin{equation}
\mathcal{S}_{\text{pop}, 3} = 
\frac{1}{NS}
\sum_{i=1}^{N}\sum_{s=1}^{S} 
\mathcal{I}\left(|c_{(i)} - c^s_{(i)}| < \epsilon\right),
\label{feature:anr}
\end{equation}
where $\epsilon$ is a small tolerance threshold, and here $ \epsilon = 1 $.
A large ANR implies a rugged landscape full of neutral moves.

\subsubsection{Non-improvable Ratio (NI)}
An individual is considered non-improvable if its sampled costs never outperform
the current value:
\begin{equation}
\alpha_i = 
\begin{cases}
1, & \text{if } \sum_{s=1}^S \mathcal{I}(c^s_{(i)} < c_{(i)}) = 0,\\
0, & \text{otherwise}.
\end{cases}
\label{feature:ni-alpha}
\end{equation}
The non-improvable ratio is then
\begin{equation}
\mathcal{S}_{\text{pop}, 4} = \frac{1}{N}\sum_{i=1}^N \alpha_i = \frac{1}{N}\sum_{i=1}^N \mathcal{I}(\sum_{s=1}^S \mathcal{I}(c^s_{(i)} < c_{(i)}) = 0).
\label{feature:ni}
\end{equation}
This indicates how many individuals are already locally optimal within the sampled
neighborhood.

\subsubsection{Non-worsenable Ratio (NW)}
Similarly, an individual is non-worsenable if not all sampled versions are strictly worse than the current cost:
\begin{equation}
\beta_i = 
\begin{cases}
1, & \text{if } \sum_{s=1}^S \mathcal{I}(c^s_{(i)} > c_{(i)}) < S,\\
0, & \text{otherwise}.
\end{cases}
\label{feature:nw-beta}
\end{equation}
The feature is
\begin{equation}
\mathcal{S}_{\text{pop}, 5} = \frac{1}{N}\sum_{i=1}^N \beta_i = \frac{1}{N}\sum_{i=1}^N \mathcal{I}(\sum_{s=1}^S \mathcal{I}(c^s_{(i)} > c_{(i)}) < S).
\label{feature:nw}
\end{equation}
A larger value implies high potential to deteriorate, reflecting evolvability.

% ============================================================

\subsection{Optimization-Progress Features}
\label{app:feature}

\subsubsection{Evaluation Progress}
Let $\text{FEs}$ be the number of consumed function evaluations and 
$\text{MaxFEs}$ the maximum budget. The progress feature is:
\begin{equation}
\mathcal{S}_{\text{progress},1} = \frac{\text{FEs}}{\text{MaxFEs}}.
\label{feature:progress}
\end{equation}

\subsubsection{Normalized Best-Cost Feature}
To represent the relative quality of the current best solution, we reuse the same normalization pattern introduced in the reward design. Let $c^{*}_{t}$ be the best cost at iteration $t$. Following the same offset transformation as reward(Section~\ref{sec:reward}), we define offset:
\begin{equation}
\delta_{\mathrm{f}}=\operatorname*{max}\bigl(1.5-c_t^{*},\,1.5-c_0^*,\,0\bigr).
\end{equation}
The normalized log-cost feature is then computed as:
\begin{equation}
\mathcal{S}_{\text{progress}, 2}=\left(\frac{\log_{10}(c_t^{*} + \delta_{\mathrm{f}})}{\log_{10}({c}_0^* + \delta_{\mathrm{f}})}\right)^{2},\qquad \mathcal{S}_{\text{progress}, 2}\in[0,1].
\end{equation}
This feature reflects the relative quality of the best solution in a scale-free manner while remaining consistent with the reward formulation. The squaring operation enhances sensitivity when $c^{*}_{t}$ approaches $c_0^*$, enabling effective reflection for concentrated values (e.g., on the Ackley problem).

\subsubsection{Global-Best Improvement Feature}

Following the same scaling principle, we define a 
progress indicator based on the relative change of consecutive global-best costs. 
Since the raw ratio $c_t^\ast / c_{t-1}^\ast$ is typically close to~1 in later stages 
of optimization, we apply an exponent $\gamma>1$ to enhance resolution near the 
upper boundary. The feature is defined as
\begin{equation}
\mathcal{S}_{\text{progress}, 3}
=
\left(\frac{c_t^\ast}{c_{t-1}^\ast}\right)^{8},
\label{feature:gb-improve}
\end{equation}
where the exponent stretches the dynamic range, from 
$\left[0.9,1\right]$ into $\left[0.43,1\right]$, for instance, making small regressions or 
stagnations more distinguishable for the policy.

% 附录A.1.3 全局最优改进特征（供对比参考）

% 附录A.1.4 分组最优改进特征（已修正）
\subsubsection{Group-Best Improvement Feature}
At the subproblem level, we use the same construction to measure the relative improvement of the group-specific best cost. 
Let $c_{t}^{*(k)}$ denote the group-best cost for subproblem $k$ at decision step $t$. The feature is computed as:
\begin{equation}
\mathcal{S}_{\text{progress}, 4} = \left(\frac{c_{t}^{*(k)}}{c_{t-1}^{*(k)}}\right)^{8},
\end{equation}
providing a fine-grained indicator of whether the decomposition-based search is progressing or stagnating within each group, which can further indicate the contribution of each subproblem.

\subsubsection{Optimizer-wise Indicators}

For each optimizer $o_l \in \mathcal{O}$ (where $l=1,\ldots,L$), we construct two 
complementary features to capture usage and effectiveness.

First, the \emph{usage feature} reflects how much computational budget 
(quantified by function evaluations, FEs) has been consumed by optimizer~$l$:
\begin{equation}
\mathcal{S}_{\text{progress}, 5}
=
\frac{\mathrm{FEs}_l}{\mathrm{FEs}_{\mathrm{total}}}.
\label{feature:opt-usage}
\end{equation}
Second, the \emph{effectiveness feature} measures the optimizer's contribution 
to global log-scale improvement. Following the stabilized denominator introduced 
in the reward design, we define
\begin{equation}
\mathcal{S}_{\text{progress}, 6}
=
\frac{\Delta\log_{10}c^{(l)}}%
     {\max\!\left(\log_{10}(c_{0}^*) 
                  - \log_{10}(c_t^{*}),\; 0.1\right)},
\label{feature:opt-effect}
\end{equation}
where $\Delta\log_{10}c^{(l)}$ denotes the optimizer-specific improvement 
contributed by $o_l$ during its execution. 
These features allow the agent to identify which optimizer remains effective 
under the current landscape and dynamically adjust its selection strategy.

\section{Benchmark Design in Detail}
\label{app:benchmark_details}

In this section, we provide a comprehensive description of the Auto-H-LSGO benchmark.

\subsection{Transformation Function}
\label{sec:transformation}
To increase the complexity of the landscape and prevent the use of simple exploitation techniques, several transformation functions are applied to the decision variables before evaluating the basic functions.

\textbf{Shift, Permutation and Rotation:}
To relocate the global optimum, remove the bias induced by variable ordering, and enforce non-separability within the subproblem's dimensions, a series of transformations is available to the decision variables, and each of them can be arbitrarily adopted or not.
If all adopted, the variables are first shifted by the global optimum $\mathbf{x}_{opt}$, followed by a permutation and a rotation:
\begin{equation}
    \mathbf{z} = \mathcal{R}\big(\xi(\mathbf{x} - \mathbf{x}_{opt})\big),
\end{equation}
where $\xi$ denotes a fixed random permutation operator on the variable indices, and a rotation matrix $\mathcal{R}$ is employed to enforce non-separability within the subproblem's dimensions.

\textbf{Oscillated Transformation ($T_{osz}$):}  
    This transformation introduces smooth local irregularities:
\begin{equation}
\begin{aligned}
T_{osz}: & \mathbb{R}^D \rightarrow \mathbb{R}^D,\\
z_i \mapsto {}  \operatorname{sign}(z_i)\exp\Big( \hat{z}_i
 +& 0.049\big(\sin(c_1\hat{z}_i)+\sin(c_2\hat{z}_i)\big) \Big),\\
 i=1, & \dots,D .
\end{aligned}
\end{equation}

    where
    \begin{equation}
        \hat{z}_i =
        \begin{cases}
            \log(|z_i|), & z_i \neq 0,\\
            0, & z_i = 0,
        \end{cases}
        \qquad
        \operatorname{sign}(z_i)=
        \begin{cases}
            -1, & z_i<0,\\
            0, & z_i=0,\\
            1, & z_i>0.
        \end{cases}
    \end{equation}
    and
    \begin{equation}
        c_1 =
        \begin{cases}
            10, & z_i>0,\\
            5.5, & \text{otherwise},
        \end{cases}
        \qquad
        c_2 =
        \begin{cases}
            7.9, & z_i>0,\\
            3.1, & \text{otherwise}.
        \end{cases}
    \end{equation}

\textbf{Asymmetry Transformation ($T_{asy}$):}  
    This transformation breaks symmetry:
    \begin{equation}
    \begin{aligned}
        T_{asy}: & \mathbb{R}^D \rightarrow \mathbb{R}^D,\\
        z_i \mapsto &
        \begin{cases}
            z_i^{\,1+\beta \frac{i-1}{D-1}\sqrt{z_i}}, & z_i>0,\\
            z_i, & \text{otherwise},
        \end{cases}\\
        i=1, & \dots,D.
    \end{aligned}
    \end{equation}
    where $\beta$ is a constant (e.g., $\beta=0.2$).

\textbf{Diagonal Scaling Matrix ($\Lambda$):}  
    A diagonal matrix $\Lambda$ is defined by
    \begin{equation}
        \lambda_i = \alpha^{\frac{1}{2}\frac{i-1}{D-1}},
        \quad i=1,\dots,D,
    \end{equation}
    and the scaling is applied as
    \begin{equation}
        z_i \leftarrow \lambda_i z_i,
    \end{equation}
    where $\alpha$ controls the condition number (e.g., $\alpha=10$).

\subsection{Basic Functions}
\label{sec:basic-function}
We select 7 basic functions to construct the test suite. Each sub-subsection below represents a basic function used in our benchmark.

\subsubsection{Sphere Function}
A smooth, unimodal function used to test basic exploitation capabilities.
\begin{equation}
\begin{aligned}
    f_1(\mathbf{z}) &= \sum_{i=1}^{D} (z'_i)^2, \\
    \text{where } \mathbf{z'} &= \Lambda T_{asy}(T_{osz}(\mathbf{z})).
\end{aligned}
\end{equation}

\subsubsection{Elliptic Function}
A high-conditioned unimodal function that creates an elongated elliptical landscape.
\begin{equation}
\begin{aligned}
    f_2(\mathbf{z}) &= \sum_{i=1}^{D} 10^{6 \frac{i-1}{D-1}} (z'_i)^2, \\
    \text{where } \mathbf{z'} &= T_{osz}(\mathbf{z}).
\end{aligned}
\end{equation}

\subsubsection{Rastrigin Function}
A complex multimodal function with a large number of local minima, testing the ability to escape local optima.
\begin{equation}
\begin{aligned}
    f_3(\mathbf{z}) &= \sum_{i=1}^{D} \left( (z'_i)^2 - 10 \cos(2\pi z'_i) + 10 \right), \\
    \text{where } \mathbf{z'} &= \Lambda T_{asy}(T_{osz}(\mathbf{z})).
\end{aligned}
\end{equation}

\subsubsection{Ackley Function}
A multimodal function with a nearly flat outer region and a steep central peak.
\begin{equation}
\begin{aligned}
    f_4(\mathbf{z}) &= -20 \exp \left( -0.2 \sqrt{\frac{1}{D} \sum_{i=1}^{D} (z'_i)^2} \right) \\
    &- \exp \left( \frac{1}{D} \sum_{i=1}^{D} \cos(2\pi z'_i) \right) + 20 + e, \\
    \text{where } \mathbf{z'} &= \Lambda T_{asy}(T_{osz}(\mathbf{z})).
\end{aligned}
\end{equation}

\subsubsection{Schwefel's Problem 1.2}
A unimodal function where variables exhibit strong dependency through cumulative sums.
\begin{equation}
\begin{aligned}
    f_5(\mathbf{z}) &= \sum_{i=1}^{D} \left( \sum_{j=1}^{i} z'_j \right)^2, \\
    \text{where } \mathbf{z'} &= T_{asy}(T_{osz}(\mathbf{z})).
\end{aligned}
\end{equation}

\subsubsection{Katsuura Function}
A highly non-smooth, fractal-like multimodal function that is extremely difficult to optimize.
\begin{equation}
\begin{aligned}
    f_6(\mathbf{z}) &= \prod_{i=1}^{D} \left( 1 + i \sum_{j=1}^{32} \frac{|2^j z'_i - \text{round}(2^j z'_i)|}{2^j} \right)^{\frac{10}{D^{1.2}}}, \\
    \text{where } \mathbf{z'} &= \Lambda T_{asy}(T_{osz}(\mathbf{z})).
\end{aligned}
\end{equation}

\subsubsection{Attractive Sector Function}
A function with highly asymmetrical slopes in different regions of the search space.
\begin{equation}
\begin{aligned}
    f_7(\mathbf{z}) &= \sum_{i=1}^{D} \phi(z'_i), \\
    \text{where } \phi(z'_i) &= \begin{cases} 
    100(z'_i)^2 + (z'_i)^4 & \text{if } z'_i > 0, \\
    (z'_i)^2 + 100(z'_i)^4 & \text{if } z'_i \leq 0,
    \end{cases} \\
    \mathbf{z'} &= \Lambda T_{asy}(T_{osz}(\mathbf{z})).
\end{aligned}
\end{equation}

\subsection{Problem Instance Configurations}
\label{sec:instance-detail}
As introduced in Section~\ref{sec:benchmark}, we constructed 18 problem instances for experiments, with their names from $\text{Ackley}_{1}$ to $\text{Ackley}_{5}$, from $\text{Attractive Sector}_{1}$ to $\text{Attractive Sector}_{5}$, and from $\text{He}_{1}$ to $\text{He}_{8}$. Here we demonstrate their configurations in detail. Namely, for a problem instance in the form of $\text{\{Basic Function\}}_{\{\text{Separability Degree}\}} =  \sum_{k=1}^K w_k f_{\text{basic}}(\mathbf{z}^{(k)})$, $\text{\{Basic Function\}}$ part indicates the basic function, which will be introduced in Section~\ref{sec:basic-configuration}, and $_{\{\text{Separability Degree}\}}$ part indicates the separability, which defines $z^{(k)}$, and it will be introduced in Section~\ref{sec:sepa-configuration}. In our problem instances we use \(K=14\) subproblems. The dimensions of these 14 subproblems are configured as follows:
two subproblems of dimension 25, three of dimension 50, three of dimension 100, two of dimension 200, two of dimension 300, one of dimension 500, and one of dimension 1000. 
In shorthand this is \(25\times2,\;50\times3,\;100\times3,\;200\times2,\;300\times2,\;500\times1,\;1000\times1\), which sums to a total problem dimension of \(3000\).

\subsubsection{Basic Function Configuration}
\label{sec:basic-configuration}
For homogeneous instances, their basic functions are indicated by their naming. For example:
\begin{equation}
    \text{Ackley} =  \sum_{k=1}^K w_k f_{4}(\mathbf{z}^{(k)}),
\end{equation}
And for heterogeneous instances, the definition is:
\begin{equation}
    \text{He} =  \sum_{k=1}^K w_k f_{\varphi(k)}(\mathbf{z}^{(k)}),
\end{equation}
where $\varphi(k)\in\{1,\dots,7\}$ is a fixed random mapping from $k$ to actual basic function, which is, in our instance, $K=14$, and 
\begin{equation}
\varphi(1\!:\!K)=\{6,2,3,1,6,7,6,7,3,4,2,5,3,2\}.
\end{equation}

\subsubsection{Separability Configuration}
\label{sec:sepa-configuration}
We introduce additional complexity to problem instances by applying transformation functions introduced in Section~\ref{sec:transformation}, controlling what $z^{(k)}$ the basic function will receive. It should be noted that the use of $T_{osz}, T_{asy}, \Lambda$ is already clarified in basic function definition part (Section~\ref{sec:basic-function}), hence we only discuss configuration including rotation matrix and overlap structure here. In the Separability Degree part, $1$ indicates that the rotation matrix is not adopted, that is
\begin{equation}
        z^{(k)} =
        \begin{cases}
            \big(\xi(\mathbf{x}^{(k)} - \mathbf{x}_{opt}^{(k)})\big), & \text{Separability Degree} = 1,\\
            \mathcal{R}\big(\xi(\mathbf{x}^{(k)} - \mathbf{x}_{opt}^{(k)})\big), & \text{otherwise}.
        \end{cases}
\end{equation}
And $2$ represents a simple configuration with rotation matrix.
A number larger than $2$ indicates a case where overlap exists within subproblems, and permutation $\xi$ is responsible for this overlap. While a normal permutation makes an injective mapping, a permutation with overlap derives common variables between adjacent subproblems, which is also known as overlap variables. The number of overlap variables between subproblem $k$ and $k+1$ is computed as $\left\lfloor \text{overlap ratio} \times \min(D_k,D_{k+1}) \right\rfloor$, which tells that larger overlap ratio comes with a more complex environment. The Separability Degree which is larger than $2$ is designed to configure such complexity, that is, $3$, $4$, $5$ represent an overlap ratio of $0.2$, $0.4$, $0.6$ respectively.

%% file: ref.bib
@article{omidvar2021review,
  title={A review of population-based metaheuristics for large-scale black-box global optimization—Part I},
  author={Omidvar, Mohammad Nabi and Li, Xiaodong and Yao, Xin},
  journal={IEEE Transactions on Evolutionary Computation},
  volume={26},
  number={5},
  pages={802--822},
  year={2021},
  publisher={IEEE}
}

@article{yao2025coevolutionary,
  title={Coevolutionary Computation and Its Applications},
  author={Yao, Xin and Chong, Siang Yew},
  year={2025},
  publisher={Springer}
}

@article{ma2018survey,
  title={A survey on cooperative co-evolutionary algorithms},
  author={Ma, Xiaoliang and Li, Xiaodong and Zhang, Qingfu and Tang, Ke and Liang, Zhengping and Xie, Weixin and Zhu, Zexuan},
  journal={IEEE Transactions on Evolutionary Computation},
  volume={23},
  number={3},
  pages={421--441},
  year={2018},
  publisher={IEEE}
}

@article{liu2024large,
  title={Large-scale evolutionary optimization: A review and comparative study},
  author={Liu, Jing and Sarker, Ruhul and Elsayed, Saber and Essam, Daryl and Siswanto, Nurhadi},
  journal={Swarm and Evolutionary Computation},
  volume={85},
  pages={101466},
  year={2024},
  publisher={Elsevier}
}

@article{li2013benchmark,
  title={Benchmark functions for the CEC 2013 special session and competition on large-scale global optimization},
  author={Li, Xiaodong and Tang, Ke and Omidvar, Mohammad N and Yang, Zhenyu and Qin, Kai and China, Hefei},
  journal={gene},
  volume={7},
  number={33},
  pages={8},
  year={2013}
}

@article{kumar2022efficient,
  title={An efficient differential grouping algorithm for large-scale global optimization},
  author={Kumar, Abhishek and Das, Swagatam and Mallipeddi, Rammohan},
  journal={IEEE Transactions on Evolutionary Computation},
  volume={28},
  number={1},
  pages={32--46},
  year={2022},
  publisher={IEEE}
}

@inproceedings{akimoto2016projection,
  title={Projection-based restricted covariance matrix adaptation for high dimension},
  author={Akimoto, Youhei and Hansen, Nikolaus},
  booktitle={Proceedings of the Genetic and Evolutionary Computation Conference 2016},
  pages={197--204},
  year={2016}
}

@article{he2020mmes,
  title={MMES: Mixture model-based evolution strategy for large-scale optimization},
  author={He, Xiaoyu and Zheng, Zibin and Zhou, Yuren},
  journal={IEEE Transactions on Evolutionary Computation},
  volume={25},
  number={2},
  pages={320--333},
  year={2020},
  publisher={IEEE}
}

@article{qiu2025novel,
  title={A Novel Two-Phase Cooperative Co-evolution Framework for Large-Scale Global Optimization with Complex Overlapping},
  author={Qiu, Wenjie and Guo, Hongshu and Ma, Zeyuan and Gong, Yue-Jiao},
  journal={arXiv preprint arXiv:2503.21797},
  year={2025}
}

@article{ma2025toward,
  title={Toward automated algorithm design: A survey and practical guide to meta-black-box-optimization},
  author={Ma, Zeyuan and Guo, Hongshu and Gong, Yue-Jiao and Zhang, Jun and Tan, Kay Chen},
  journal={IEEE Transactions on Evolutionary Computation},
  year={2025},
  publisher={IEEE}
}

@article{ma2025metabox,
  title={MetaBox-v2: A Unified Benchmark Platform for Meta-Black-Box Optimization},
  author={Ma, Zeyuan and Gong, Yue-Jiao and Guo, Hongshu and Qiu, Wenjie and Ma, Sijie and Lian, Hongqiao and Zhan, Jiajun and Chen, Kaixu and Wang, Chen and Huang, Zhiyang and others},
  journal={arXiv preprint arXiv:2505.17745},
  year={2025}
}

@article{guo2025designx,
  title={DesignX: Human-Competitive Algorithm Designer for Black-Box Optimization},
  author={Guo, Hongshu and Ma, Zeyuan and Ma, Yining and Zhang, Xinglin and Chen, Wei-Neng and Gong, Yue-Jiao},
  journal={arXiv preprint arXiv:2505.17866},
  year={2025}
}

@inproceedings{auger2012tutorial,
  title={Tutorial CMA-ES: evolution strategies and covariance matrix adaptation},
  author={Auger, Anne and Hansen, Nikolaus},
  booktitle={Proceedings of the 14th annual conference companion on Genetic and evolutionary computation},
  pages={827--848},
  year={2012}
}

@inproceedings{ros2008simple,
  title={A simple modification in CMA-ES achieving linear time and space complexity},
  author={Ros, Raymond and Hansen, Nikolaus},
  booktitle={International conference on parallel problem solving from nature},
  pages={296--305},
  year={2008},
  organization={Springer}
}

@article{li2017simple,
  title={A simple yet efficient evolution strategy for large-scale black-box optimization},
  author={Li, Zhenhua and Zhang, Qingfu},
  journal={IEEE Transactions on Evolutionary Computation},
  volume={22},
  number={5},
  pages={637--646},
  year={2017},
  publisher={IEEE}
}

@article{li2018fast,
  title={Fast covariance matrix adaptation for large-scale black-box optimization},
  author={Li, Zhenhua and Zhang, Qingfu and Lin, Xi and Zhen, Hui-Ling},
  journal={IEEE transactions on cybernetics},
  volume={50},
  number={5},
  pages={2073--2083},
  year={2018},
  publisher={IEEE}
}

@article{guo2025advancing,
  title={Advancing cma-es with learning-based cooperative coevolution for scalable optimization},
  author={Guo, Hongshu and Qiu, Wenjie and Ma, Zeyuan and Zhang, Xinglin and Zhang, Jun and Gong, Yue-Jiao},
  journal={arXiv preprint arXiv:2504.17578},
  year={2025}
}

@inproceedings{ma2025accurate,
  title={Accurate peak detection in multimodal optimization via approximated landscape learning},
  author={Ma, Zeyuan and Lian, Hongqiao and Qiu, Wenjie and Gong, Yue-Jiao},
  booktitle={Proceedings of the Genetic and Evolutionary Computation Conference},
  pages={1127--1136},
  year={2025}
}

@inproceedings{cenikj2024transoptas,
  title={Transoptas: Transformer-based algorithm selection for single-objective optimization},
  author={Cenikj, Gjorgjina and Petelin, Ga{\v{s}}per and Eftimov, Tome},
  booktitle={Proceedings of the Genetic and Evolutionary Computation Conference Companion},
  pages={403--406},
  year={2024}
}

@article{cenikj2024survey,
  title={A survey of meta-features used for automated selection of algorithms for black-box single-objective continuous optimization},
  author={Cenikj, Gjorgjina and Nikolikj, Ana and Petelin, Ga{\v{s}}per and van Stein, Niki and Doerr, Carola and Eftimov, Tome},
  journal={arXiv preprint arXiv:2406.06629},
  year={2024}
}

@article{schulman2017proximal,
  title={Proximal policy optimization algorithms},
  author={Schulman, John and Wolski, Filip and Dhariwal, Prafulla and Radford, Alec and Klimov, Oleg},
  journal={arXiv preprint arXiv:1707.06347},
  year={2017}
}

@article{duan2024pypop7,
  title={Pypop7: A pure-python library for population-based black-box optimization},
  author={Duan, Qiqi and Zhou, Guochen and Shao, Chang and Wang, Zhuowei and Feng, Mingyang and Huang, Yuwei and Tan, Yajing and Yang, Yijun and Zhao, Qi and Shi, Yuhui},
  journal={Journal of Machine Learning Research},
  volume={25},
  number={296},
  pages={1--28},
  year={2024}
}

@inproceedings{lange2023evosax,
  title={evosax: Jax-based evolution strategies},
  author={Lange, Robert Tjarko},
  booktitle={Proceedings of the Companion Conference on Genetic and Evolutionary Computation},
  pages={659--662},
  year={2023}
}

@article{omidvar2017dg2,
  title={DG2: A faster and more accurate differential grouping for large-scale black-box optimization},
  author={Omidvar, Mohammad Nabi and Yang, Ming and Mei, Yi and Li, Xiaodong and Yao, Xin},
  journal={IEEE Transactions on Evolutionary Computation},
  volume={21},
  number={6},
  pages={929--942},
  year={2017},
  publisher={IEEE}
}

@article{liu2025evolutionary,
  title={Evolutionary Contribution and Problem Heuristic Information Ensemble-Based Resource Allocation for Cooperative Coevolution},
  author={Liu, Dong and Lu, Ming-Yuan and Yang, Qiang and Chen, Wei-Neng and Jia, Ya-Hui and Li, Jian-Yu and Li, Tao and Ma, Yuan-Yuan and Zhang, Jun},
  journal={IEEE Transactions on Evolutionary Computation},
  year={2025},
  publisher={IEEE}
}

@inproceedings{komarnicki2024overlapping,
  title={Overlapping Cooperative Co-Evolution for Overlapping Large-Scale Global Optimization Problems},
  author={Komarnicki, Marcin Michal and Przewozniczek, Michal Witold and Tin{\'o}s, Renato and Li, Xiaodong},
  booktitle={Proceedings of the Genetic and Evolutionary Computation Conference},
  pages={665--673},
  year={2024}
}

@article{liang2006comprehensive,
  title={Comprehensive learning particle swarm optimizer for global optimization of multimodal functions},
  author={Liang, Jing J and Qin, A Kai and Suganthan, Ponnuthurai N and Baskar, S},
  journal={IEEE transactions on evolutionary computation},
  volume={10},
  number={3},
  pages={281--295},
  year={2006},
  publisher={IEEE}
}

@article{garcia2008global,
  title={Global and local real-coded genetic algorithms based on parent-centric crossover operators},
  author={Garc{\'\i}a-Mart{\'\i}nez, Carlos and Lozano, Manuel and Herrera, Francisco and Molina, Daniel and S{\'a}nchez, Ana M},
  journal={European journal of operational research},
  volume={185},
  number={3},
  pages={1088--1113},
  year={2008},
  publisher={Elsevier}
}

@incollection{yao2025cooperative,
  title={Cooperative Coevolution for Large-Scale Optimization},
  author={Yao, Xin and Chong, Siang Yew},
  booktitle={Coevolutionary Computation and Its Applications},
  pages={199--270},
  year={2025},
  publisher={Springer}
}

@inproceedings{ha2019large,
  title={Large-scale design-economics optimization of eVTOL concepts for urban air mobility},
  author={Ha, Tae H and Lee, Keunseok and Hwang, John T},
  booktitle={AIAA Scitech 2019 Forum},
  pages={1218},
  year={2019}
}

@article{thomas2017improving,
  title={Improving the FLORIS wind plant model for compatibility with gradient-based optimization},
  author={Thomas, Jared J and Gebraad, Pieter MO and Ning, Andrew},
  journal={Wind Engineering},
  volume={41},
  number={5},
  pages={313--329},
  year={2017},
  publisher={SAGE Publications Sage UK: London, England}
}

@article{galvan2021neuroevolution,
  title={Neuroevolution in deep neural networks: Current trends and future challenges},
  author={Galv{\'a}n, Edgar and Mooney, Peter},
  journal={IEEE Transactions on Artificial Intelligence},
  volume={2},
  number={6},
  pages={476--493},
  year={2021},
  publisher={IEEE}
}

@article{ibrahimov2012evolutionary,
  title={Evolutionary approaches for supply chain optimisation. Part II: multi-silo supply chains},
  author={Ibrahimov, Maksud and Mohais, Arvind and Schellenberg, Sven and Michalewicz, Zbigniew},
  journal={International Journal of Intelligent Computing and Cybernetics},
  volume={5},
  number={4},
  pages={473--499},
  year={2012},
  publisher={Emerald Group Publishing Limited}
}

@article{song2019divide,
  title={A divide-and-conquer evolutionary algorithm for large-scale virtual network embedding},
  author={Song, An and Chen, Wei-Neng and Gong, Yue-Jiao and Luo, Xiaonan and Zhang, Jun},
  journal={IEEE Transactions on Evolutionary Computation},
  volume={24},
  number={3},
  pages={566--580},
  year={2019},
  publisher={IEEE}
}

@article{hwang2014large,
  title={Large-scale multidisciplinary optimization of a small satellite’s design and operation},
  author={Hwang, John T and Lee, Dae Young and Cutler, James W and Martins, Joaquim RRA},
  journal={Journal of Spacecraft and Rockets},
  volume={51},
  number={5},
  pages={1648--1663},
  year={2014},
  publisher={American Institute of Aeronautics and Astronautics}
}

@article{xu2023large,
  title={A large-scale continuous optimization benchmark suite with versatile coupled heterogeneous modules},
  author={Xu, Peilan and Luo, Wenjian and Lin, Xin and Zhang, Jiajia and Wang, Xuan},
  journal={Swarm and Evolutionary Computation},
  volume={78},
  pages={101280},
  year={2023},
  publisher={Elsevier}
}

@article{xu2025autoep,
  title={AutoEP: LLMs-Driven Automation of Hyperparameter Evolution for Metaheuristic Algorithms},
  author={Xu, Zhenxing and Zhang, Yizhe and Bao, Weidong and Wang, Hao and Chen, Ming and Ye, Haoran and Jiang, Wenzheng and Yan, Hui and Wang, Ji},
  journal={arXiv preprint arXiv:2509.23189},
  year={2025}
}

@inproceedings{stanovov2021nl,
  title={NL-SHADE-RSP algorithm with adaptive archive and selective pressure for CEC 2021 numerical optimization},
  author={Stanovov, Vladimir and Akhmedova, Shakhnaz and Semenkin, Eugene},
  booktitle={2021 IEEE Congress on Evolutionary Computation (CEC)},
  pages={809--816},
  year={2021},
  organization={IEEE}
}

@inproceedings{li2006improved,
  title={An improved particle swarm optimizer for truss structure optimization},
  author={Li, Lijuan and Huang, Zhibin and Liu, Feng},
  booktitle={2006 International Conference on Computational Intelligence and Security},
  volume={1},
  pages={924--928},
  year={2006},
  organization={IEEE}
}

@misc{du2025metablackboxoptimizationbispacelandscape,
      title={Meta-Black-Box Optimization with Bi-Space Landscape Analysis and Dual-Control Mechanism for SAEA}, 
      author={Yukun Du and Haiyue Yu and Xiaotong Xie and Yan Zheng and Lixin Zhan and Yudong Du and Chongshuang Hu and Boxuan Wang and Jiang Jiang},
      year={2025},
      eprint={2511.15551},
      archivePrefix={arXiv},
      primaryClass={cs.NE},
      url={https://arxiv.org/abs/2511.15551}, 
}

@ARTICLE{van2025llamea,
  author={Stein, Niki van and Bäck, Thomas},
  journal={IEEE Transactions on Evolutionary Computation},
  title={LLaMEA: A Large Language Model Evolutionary Algorithm for Automatically Generating Metaheuristics},
  year={2025},
  volume={29},
  number={2},
  pages={331-345},
  doi={10.1109/TEVC.2024.3497793}
}

@article{tian2024enhanced,
  title={An enhanced differential grouping method for large-scale overlapping problems},
  author={Tian, Maojiang and Chen, Mingke and Du, Wei and Tang, Yang and Jin, Yaochu},
  journal={IEEE Transactions on Evolutionary Computation},
  year={2024},
  publisher={IEEE}
}

@article{chen2025metade,
  title={Metade: Evolving differential evolution by differential evolution},
  author={Chen, Minyang and Feng, Chenchen and Cheng, Ran},
  journal={IEEE Transactions on Evolutionary Computation},
  year={2025},
  publisher={IEEE}
}

@article{song2024reinforcement,
  title={Reinforcement learning-assisted evolutionary algorithm: A survey and research opportunities},
  author={Song, Yanjie and Wu, Yutong and Guo, Yangyang and Yan, Ran and Suganthan, Ponnuthurai Nagaratnam and Zhang, Yue and Pedrycz, Witold and Das, Swagatam and Mallipeddi, Rammohan and Ajani, Oladayo Solomon and others},
  journal={Swarm and Evolutionary Computation},
  volume={86},
  pages={101517},
  year={2024},
  publisher={Elsevier}
}

@article{li2024bridging,
  title={Bridging evolutionary algorithms and reinforcement learning: A comprehensive survey on hybrid algorithms},
  author={Li, Pengyi and Hao, Jianye and Tang, Hongyao and Fu, Xian and Zhen, Yan and Tang, Ke},
  journal={IEEE Transactions on evolutionary computation},
  year={2024},
  publisher={IEEE}
}

@incollection{emmerich2025evolution,
  title={Evolution strategies},
  author={Emmerich, Michael and Shir, Ofer M and Wang, Hao},
  booktitle={Handbook of heuristics},
  pages={89--123},
  year={2025},
  publisher={Springer}
}

@article{pant2020differential,
  title={Differential Evolution: A review of more than two decades of research},
  author={Pant, Millie and Zaheer, Hira and Garcia-Hernandez, Laura and Abraham, Ajith and others},
  journal={Engineering Applications of Artificial Intelligence},
  volume={90},
  pages={103479},
  year={2020},
  publisher={Elsevier}
}

@article{wang2018particle,
  title={Particle swarm optimization algorithm: an overview},
  author={Wang, Dongshu and Tan, Dapei and Liu, Lei},
  journal={Soft computing},
  volume={22},
  number={2},
  pages={387--408},
  year={2018},
  publisher={Springer}
}

@article{qiu2026automated,
  title   = {Automated Algorithm Design for Black-Box Optimization: Progress and Challenges},
  author  = {Qiu, Wen-Jie and Guo, Hong-Shu and Ma, Ze-Yuan and Zhang, Jun and Gong, Yue-Jiao},
  journal = {Chinese Journal of Computers},
  volume  = {49},
  number  = {4},
  year    = {2026},
  month   = apr,
  doi     = {10.11897/SP.J.1016.2026.00855}
}

@article{gong2023offline,
  title={Offline data-driven optimization at scale: A cooperative coevolutionary approach},
  author={Gong, Yue-Jiao and Zhong, Yuan-Ting and Huang, Hao-Gan},
  journal={IEEE Transactions on Evolutionary Computation},
  volume={28},
  number={6},
  pages={1809--1823},
  year={2023},
  publisher={IEEE}
}
